%% file: marine_eval.tex
\documentclass{article} 
\usepackage[table]{xcolor}
\usepackage{iclr2023_conference,times}
\input{math_commands.tex}

\usepackage{textcomp}
\usepackage{diagbox}
\usepackage{hyperref}       
\usepackage{url}            
\usepackage{booktabs}       
\usepackage{amsfonts}       
\usepackage{nicefrac}       
\usepackage{microtype}      
\usepackage{bbm}          
\usepackage{xcolor}         
\usepackage{booktabs}
\usepackage{graphicx}
\usepackage{multirow}
\usepackage{wrapfig}
\usepackage{placeins}
\usepackage{float}
\usepackage{bbm}
\usepackage{listings}
\usepackage{nameref}
\usepackage{varioref}
\usepackage{fancyhdr}
\usepackage{enumitem}
\usepackage{cases}
\usepackage{xcolor}
\usepackage{ulem}
\setlist[itemize]{leftmargin=*}
\usepackage[ruled,vlined]{algorithm2e}
\definecolor{commentcolor}{RGB}{110,154,155}

\definecolor{pearThree}{HTML}{E74C3C}
\definecolor{pearDarker}{HTML}{1D2DEC}
\definecolor{darkTurquoise}{HTML}{007C7D}
\hypersetup{
	colorlinks,
	citecolor=darkTurquoise,
	linkcolor=pearThree,
	breaklinks=true,
	urlcolor=pearDarker}
\definecolor{shadecolor}{rgb}{0.92,0.92,0.92}

\title{Exploring Boundary of GPT-4V on Marine Analysis: A Preliminary Case Study}

\author{Ziqiang Zheng\textsuperscript{1}, Yiwei Chen\textsuperscript{1}, Jipeng Zhang\textsuperscript{1}, Tuan-Anh Vu\textsuperscript{1}, Huimin Zeng\textsuperscript{2}, Yue Him Wong Tim\textsuperscript{3}, Sai-Kit Yeung\textsuperscript{1} \\
\textsuperscript{1}The Hong Kong University of Science and Technology, \\
\textsuperscript{2}University of Science and Technology of China,
\textsuperscript{3}Shenzhen University\\
\small{\texttt{\{zzhengaw,ychenmb,jzhanggr,tavu\}@connect.ust.hk}, \texttt{saikit@ust.hk}} \\
}

\iclrfinalcopy 
\begin{document}
\maketitle

\begin{abstract}
Large language models (LLMs) have demonstrated a powerful ability to answer various queries as a general-purpose assistant. The continuous multi-modal large language models (MLLM) empower LLMs with the ability to perceive visual signals. The launch of GPT-4 (Generative Pre-trained Transformers) has generated significant interest in the research communities. GPT-4V(ison) has demonstrated significant power in both academia and industry fields, as a focal point in a new artificial intelligence generation. Though significant success was achieved by GPT-4V, exploring MLLMs in domain-specific analysis (\emph{e.g.}, marine analysis) that required domain-specific knowledge and expertise has gained less attention. In this study, we carry out the preliminary and comprehensive case study of utilizing GPT-4V for marine analysis. This report conducts a systematic evaluation of existing GPT-4V, assessing the performance of GPT-4V on marine research and also setting a new standard for future developments in MLLMs. The experimental results of GPT-4V show that the responses generated by GPT-4V are still far away from satisfying the domain-specific requirements of the marine professions. All images and prompts used in this study will be available at \textcolor{blue}{\href{https://github.com/hkust-vgd/Marine\_GPT-4V\_Eval}{https://github.com/hkust-vgd/Marine\_GPT-4V\_Eval}}
\end{abstract}

\thispagestyle{empty}

{
  \hypersetup{linkcolor=black}
  \tableofcontents
}

\newpage

{
  \hypersetup{linkcolor=black}
  \listoffigures
}

\newpage

\pagenumbering{arabic}

\section{Introduction}
Large language models (LLMs)~\citep{raffel2020exploring,chiang2023vicuna,zhang2022opt,touvron2023llama,touvron2023llama2,ouyang2022training,openai2023gpt4,brown2020gpt3,scao2022bloom} demonstrated an impressive ability to handle a large range of user-tailored tasks. As a general-purpose assistant, ChatGPT/GPT-4~\citep{openai2023gpt4,ouyang2022training} could understand human intents and complete various real-world tasks. The development of multi-modal large language models~\citep{li2023blip,zhu2023minigpt,zheng2023marinegpt,peng2023instruction,team2023gemini,alayrac2022flamingo} (MLLMs) such as GPT-4V represents an important step towards more sophisticated AI systems with the ability to receive both textual inputs and visual data. The integration of vision in language models has marked a significant milestone. GPT-4V showcased impressive general-purpose visual understanding and reasoning abilities. The advent of GPT-4V has expanded AI applications, aligning with the multi-modal capabilities of the human brain. In detail, GPT-4V extends the abilities of GPT-4 to analyze and interpret images and has attracted significant attention across both academia and industry.

Existing open-source general-purpose MLLMs~\citep{liu2023visual,peng2023kosmos,li2023otter} often lack in image-text analysis~\citep{lu2022learn} due to limited model size and data scale. It is still unclear how GPT-4V, and MLLMs built on GPT-4, perform various multimodal understanding tasks. Though vision capabilities embodied in GPT-4 have pioneered new avenues for advanced image-text analysis, the challenges~\citep{fu2023mme,singh2023assessing} of evaluating how GPT-4V accurately perceives visual signals and measuring the effectiveness of such a system arise. To evaluate whether GPT-4V could achieve robust visual perception and mimic the inherently subjective and associative processes of human perception, recent studies~\citep{yang2023dawn,zhang2023gpt,fu2023challenger,ge2023mllm,bubeck2023sparks} have been conducted to evaluate the performance of GPT-4V in different areas, such as recommendation~\citep{zhou2023exploring}, medical analysis~\citep{li2023llava}, radiological~\citep{busch2023text}, mathematic~\citep{gao2023g}, and general-purpose visual analysis tasks~\citep{yang2023dawn,bubeck2023sparks}. Evaluating the performance of GPT-4V in these areas will provide insights into the flexibility of GPT-4V as the AI assistant. However, there are few attempts~\citep{palnitkar2023chatsim,zheng2023marinegpt} to utilize GPT-4V for more advanced analysis, which requires advanced and domain-specific knowledge and expertise. 

To bridge this gap, we present a preliminary case study investigating the marine analysis based on GPT-4V. We explore whether GPT-4V could serve as an effective visual perception system and a professional expert for sensitive, informative, and accurate knowledge delivery. We construct a series of qualitative test samples spanning multiple purposes in the field of marine analysis and employ these samples to assess the quality of the responses generated by GPT-4V.

We propose to evaluate the performance of GPT-4V on marine analysis from the following aspects: \textit{perception}, \textit{statistics}, \textit{domain-specific question answering}, \textit{marine culture understanding}, \textit{advanced functions} and \textit{prompt engineering}. We pick up images that are not accessible online or private data, combined with manually crafted prompts to build the evaluation samples. Evaluation results on our constructed testing samples prove that GPT-4V has a remarkable OCR, event detection, and framework understanding ability across various conditions, due to its robust visual-text comprehension capabilities and extensive knowledge. However, we have also observed the intrinsic limitations of using GPT-4V for marine analysis. GPT-4V only demonstrates very limited fine-grained marine object recognition ability and is easily misled by meticulously forged filenames (we observe that GPT-4V will read the filenames of uploaded images as context prompts). Besides, GPT-4V cannot perform complicated object counting and detect all the objects within the visual images since it is mainly performing image-level understanding. GPT-4V also failed to accurately capture subtle details in images and respond with the required domain-specific information. We finally demonstrate that GPT-4V cannot conduct advanced marine analysis as a professional analysis tool. We summarize our findings as follows.
\begin{itemize}
    \item In this study, we embark on an in-depth analysis of GPT-4V on domain-specific marine analysis. The expert capacity of GPT-4V has been measured for applying the learned domain knowledge and skills to the professional domains. Our study holds significant importance for the marine research community, providing valuable insights and guidance for future exploration of utilizing MLLMs for domain-specific analysis.
    \item We demonstrate several limitations of GPT-4V on marine analysis. Despite these limitations, we also aim to include a list of potential abilities of GPT-4V that we have identified as a domain-specific analysis tool. We hope that these explorations and our constructed domain-specific testing samples can offer valuable insights and serve as domain-specific benchmark data for evaluating MLLMs on domains with professional knowledge.
    \item We also acknowledge GPT-4V could be easily misled by the wrong prompts (\emph{e.g.}, the filenames of visual images), demonstrating GPT-4V leans towards the text prompts and without looking at the visual elements within the images. The hallucination happens a lot when GPT-4V is asked to answer domain-specific questions. 
\end{itemize}

\section{Experiments}
\subsection{Approach}
\noindent\textbf{Data construction}. To avoid the testing sample leakage, all the samples involved in this study are from different sources: 1) private data collection contributed by marine biologists~\citep{zheng2023marine}; 2) manually cropped frames from YouTube videos; 3) Internet images posted after the release of GPT-4V APIs; 4) framework and flowchart images from research articles and books~\citep{haixin2023marinedet,ziqiang2023coralvos}; and 5) images from public datasets~\citep{beijbom2015towards} and our newly created images. To promote the consistency and reliability of our study and increase the robustness of our findings, we make sure that every case has at least 10 testing samples with high diversity.

\noindent\textbf{Prompt design}. GPT-4V has been demonstrated to support a diverse range of visual processing based on various signed prompts~\citep{wang2022self,peng2023instruction}. This inspires us to design the various prompts. Our prompts in this study are characterized by a rich diversity and complexity of instructions to enable GPT-4V to generate comprehensive and descriptive responses, which are aligned with the user intents. 

\noindent\textbf{Evaluation metric}. In each testing case, we compute the accuracy of GPT-4V on a wide range of visual tasks. For those object recognition tasks with ground truth labeled by the domain experts, we evaluate whether GPT-4V could yield satisfactory object recognition performance according to the generated labels. For those evaluation metrics with human judgment involved, we mainly design two protocols~\citep{zhang2023gpt,ge2023mllm}: \textit{pairwise comparison} and \textit{image-based scoring}. For pairwise comparison, we judge whether the two images come from the same identity or the same species. For pairwise scoring, we ask both GPT-4V and human labelers to generate scores on a scale of 1 to 10. The ground truths under the two protocols are both generated by human experts.

\subsection{Perception} 
In this section, our goal is to assess the performance of GPT-4V in various challenging vision tasks. The involved tasks demand a powerful visual perception ability to understand the real world. Our experiments focus on the ability of GPT-4V to sense the visual contents and then perform image-level, object-level and attribute-level comprehension.

\begin{figure}[htbp]
\begin{center}
\includegraphics[width=\textwidth]{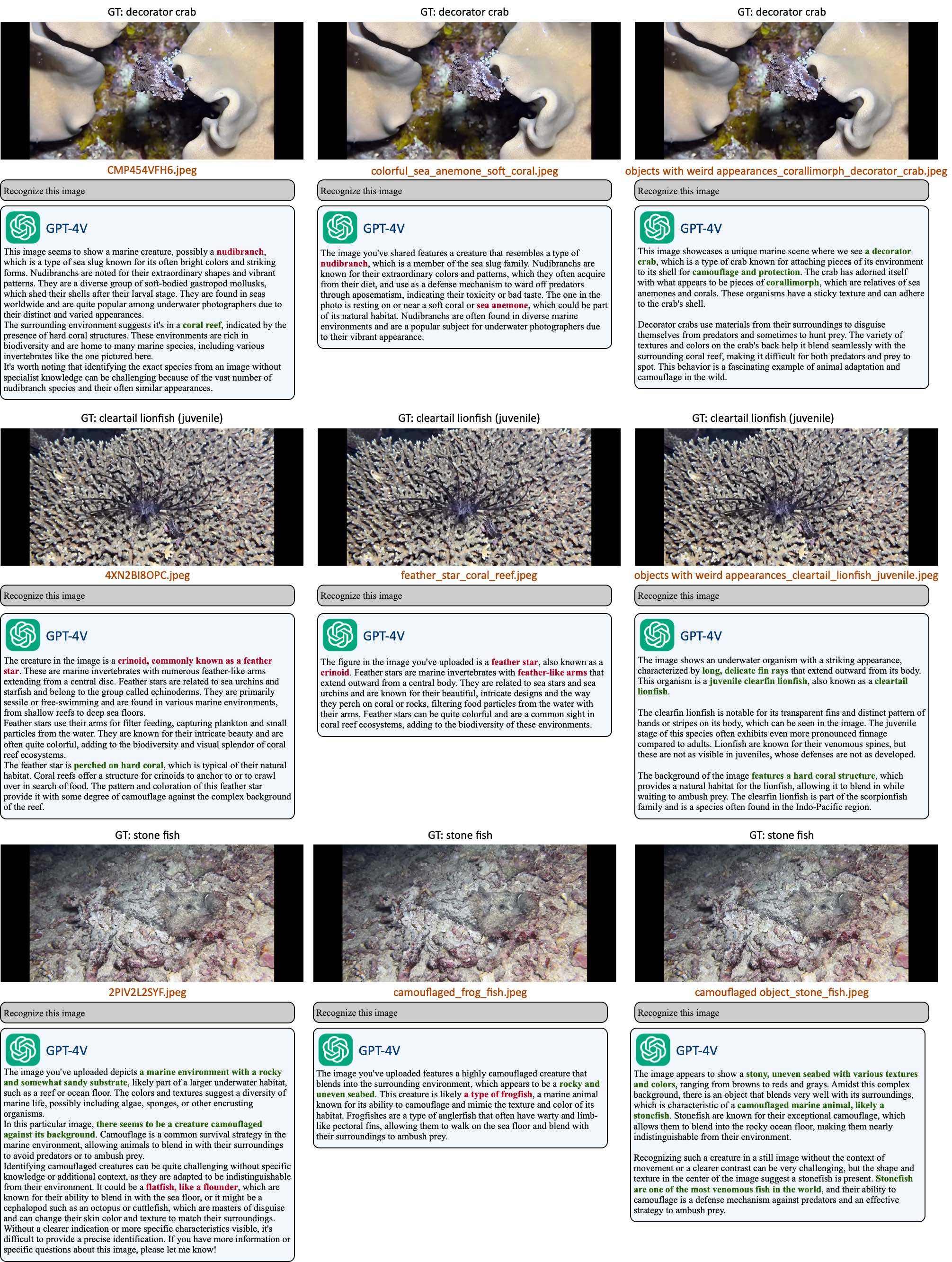}
\end{center}
\caption[Filename comparison]{The marine object recognition results under three different settings: left column (with \textbf{random} filename); middle column (with meticulously forged \textbf{misleading} filename); and right column (with \textbf{meaningful} and \textbf{aligned} filename). The texts in \textcolor{red}{red} represent the wrong responses and texts in \textcolor{green}{green} indicate the correct responses. The prompts are ``Recognize the object in this figure''.}
\label{fig:mislead}
\end{figure}

\begin{figure}[htbp]
\begin{center}
\includegraphics[width=0.85\textwidth]{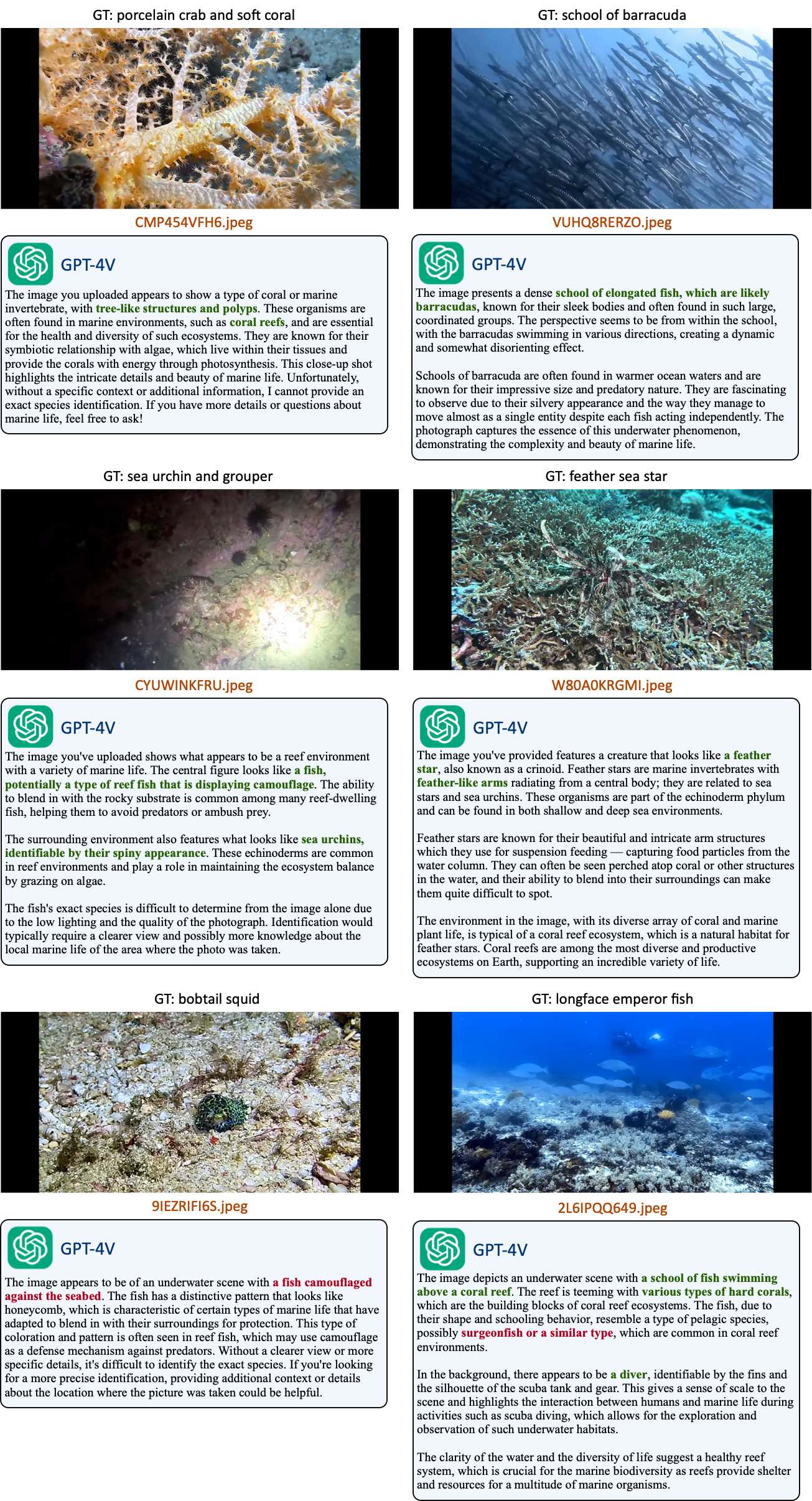}
\end{center}
\caption[Testing with random filenames]{The marine object recognition results under the setting with random filenames. The prompts are ``Recognize the object in this figure''.}
\label{fig:random}
\vspace{-0.2in}
\end{figure}

\begin{figure}[htbp]
\begin{center}
\includegraphics[width=0.9\textwidth]{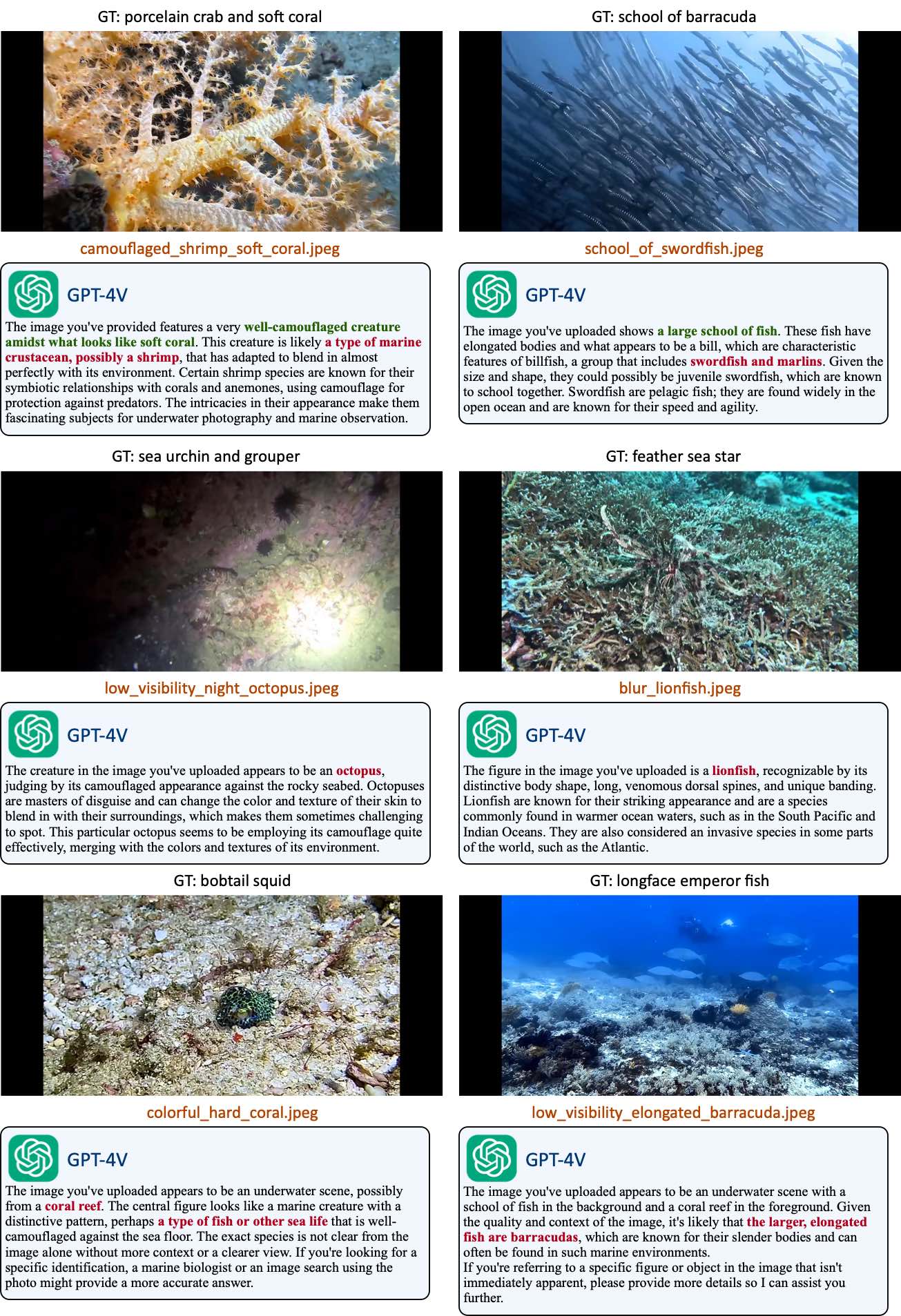}
\end{center}
\caption[Testing with misleading filenames]{The marine object recognition results under the setting with meticulously forged \textbf{misleading} filenames. The prompts are ``Recognize the object in this figure''.}
\label{fig:mislead_more}
\end{figure}

\begin{figure}[htbp]
\vspace{-0.2in}
\begin{center}
\includegraphics[width=0.88\textwidth]{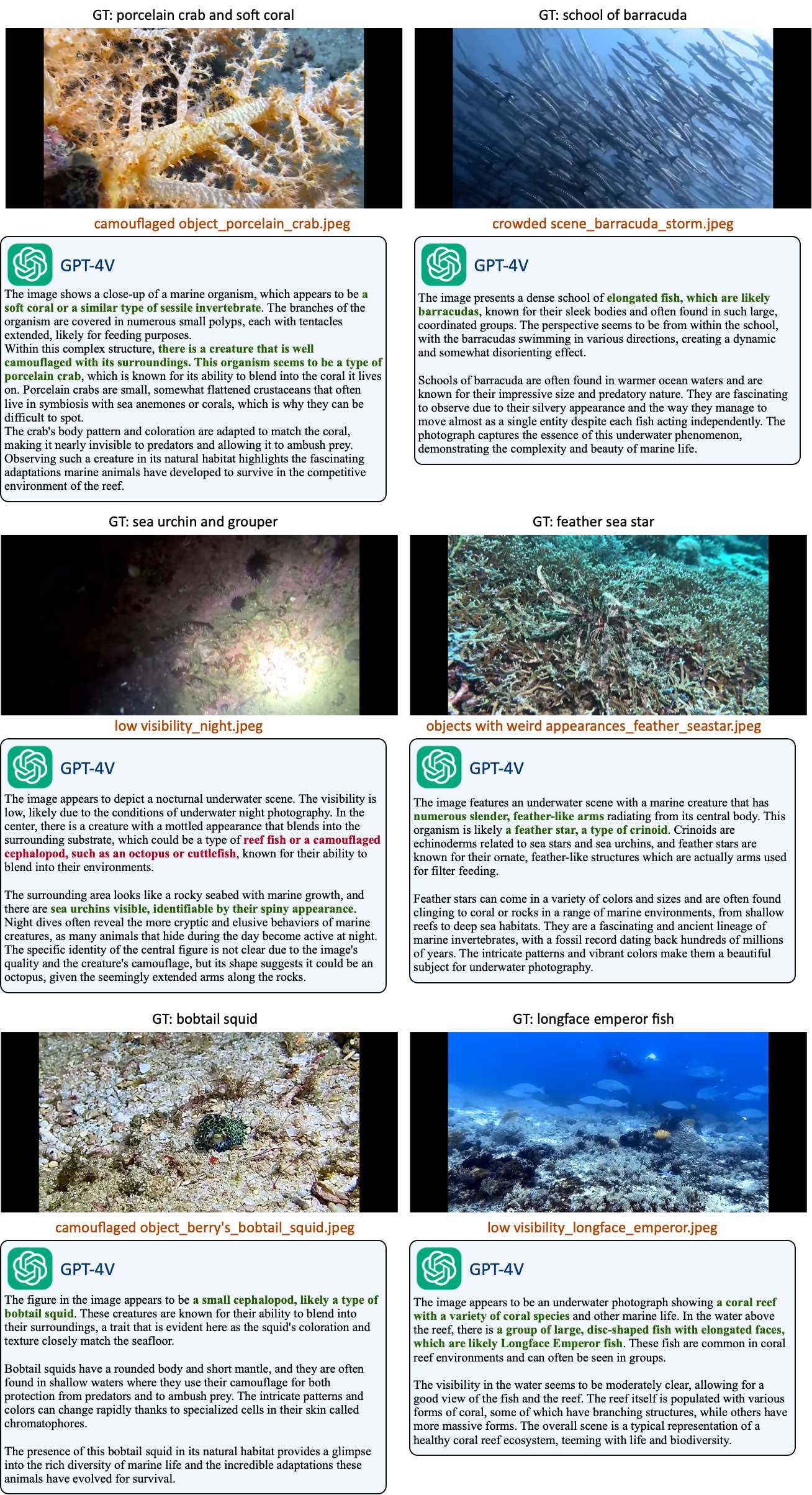}
\end{center}
\caption[Testing with meaningful filenames]{The marine object recognition results under the setting with \textbf{meaningful} and \textbf{aligned} filenames. The prompts are ``Recognize the object in this figure''.}
\label{fig:correct_name}
\vspace{-0.2in}
\end{figure}

We first explore whether GPT-4V could really understand the visual content of the given marine images or just respond without looking at the visual signals. We perform experiments using the same images under three settings: 1) with \textbf{random} filename; 2) with meticulously forged \textbf{misleading} filename; and 3) with \textbf{meaningful} and \textbf{aligned} filename. The experimental results are illustrated in Figure~\ref{fig:mislead}. The filenames and the ground truths of the marine objects are also provided as references. As illustrated, we observe that GPT-4V will recognize the marine objects within the given image under the first setting since no side clues are provided. GPT-4V tends to describe all the appeared meaningful objects and usually yields \textbf{longer} responses. Under the second setting, with the misleading filename given, GPT-4V will respond according to the given file name and generate some ``false promise'' that does \textbf{not} appear in the image. GPT-4V could be easily deceived by the meticulously forged filenames and yield some wrong answers. We guess that GPT-4V would read the filename of the uploaded image and regard such filename as the context prompt when generating the responses. It will easily produce a hallucination if the wrong context prompts do not exist in the image. As for the final setting, when the correct and aligned filenames are given, GPT-4V could generate meaningful and satisfactory responses. However, we cannot claim that GPT-4V could really understand the visual contents of uploaded images since abstracted conception names have already leaked in the filenames. More inference results under the three settings are provided in Figure~\ref{fig:random}, Figure~\ref{fig:mislead_more}, and Figure~\ref{fig:correct_name}, respectively.

\textbf{Considering the conception leakage issue, we rename all the images in all our experiments to meaningless filenames to avoid information leakage and ensure fair testing}. 

\newpage
\subsubsection{Marine object recognition}
\begin{figure}[htbp]
\begin{center}
\includegraphics[width=\textwidth]{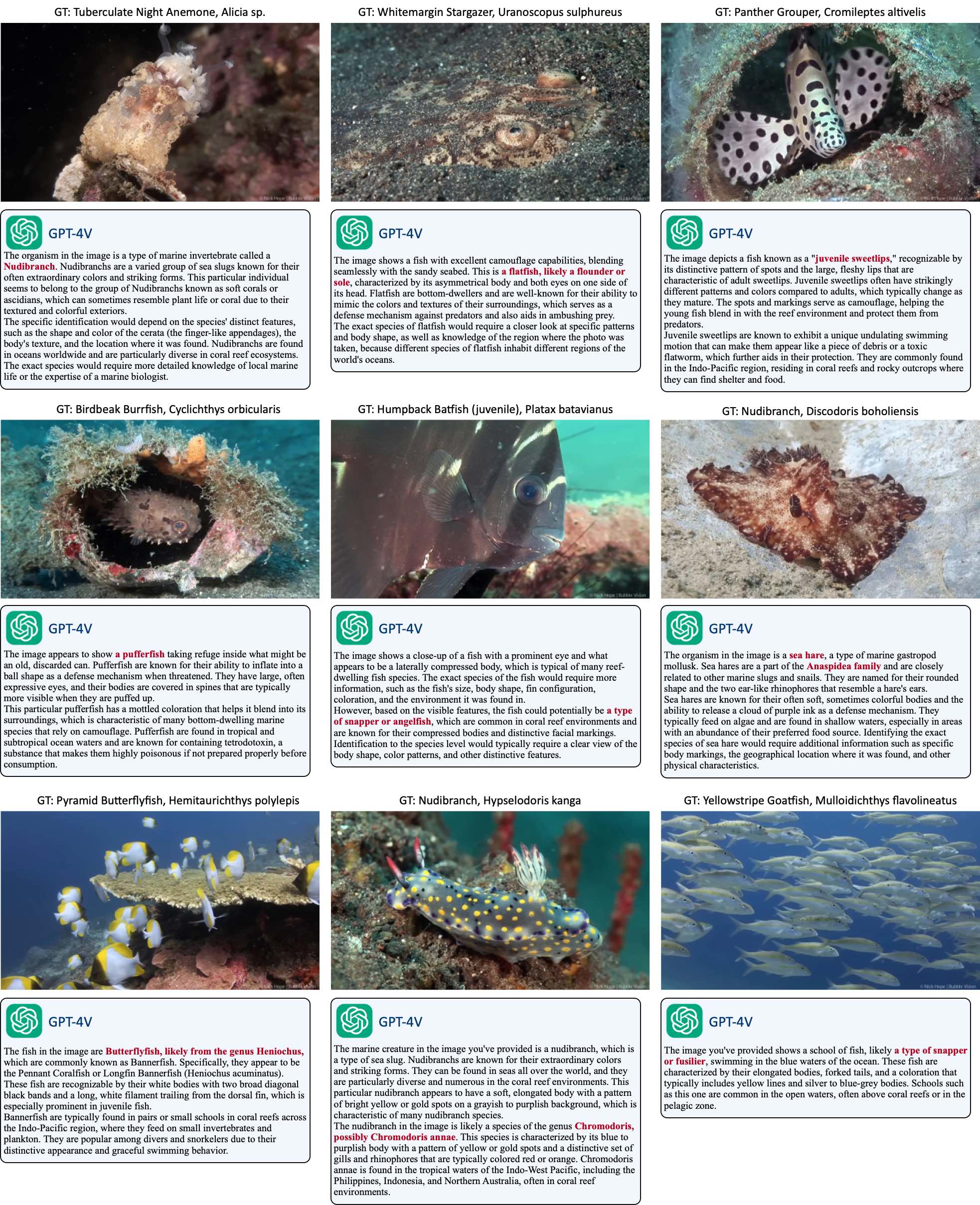}
\end{center}
\caption[Wide spectrum of marine object recognition]{The marine object recognition results of recognizing a wide spectrum of marine objects. The prompts are ``Recognize this image and tell me the species name of the recognized objects''. The ground truths are also provided.}
\label{fig:recognition_results}
\end{figure}

\noindent\textbf{Wide spectrum of marine object recognition}. We first explore whether GPT-4V could recognize a wide range of marine objects. We pick up 300 different marine images that contain the salient visual elements from one single marine species. In other words, there are 300 different marine species involved in our experiments. These images are manually cropped from the Youtube videos or the MVK dataset~\citep{truong2023marine,zheng2023marine} The ground truth of the appeared marine objects is labeled by domain experts and we manually compared the recognized object names with the ground truth for computing the recognition accuracy. Some marine object recognition results are provided in Figure~\ref{fig:recognition_results}. As illustrated, GPT-4V failed to accurately recognize marine objects that are not relatively common. There is still a very large room to improve the recognition accuracy of GPT-4V on marine object recognition.

\begin{figure}[htbp]
\begin{center}
\includegraphics[width=0.95\textwidth]{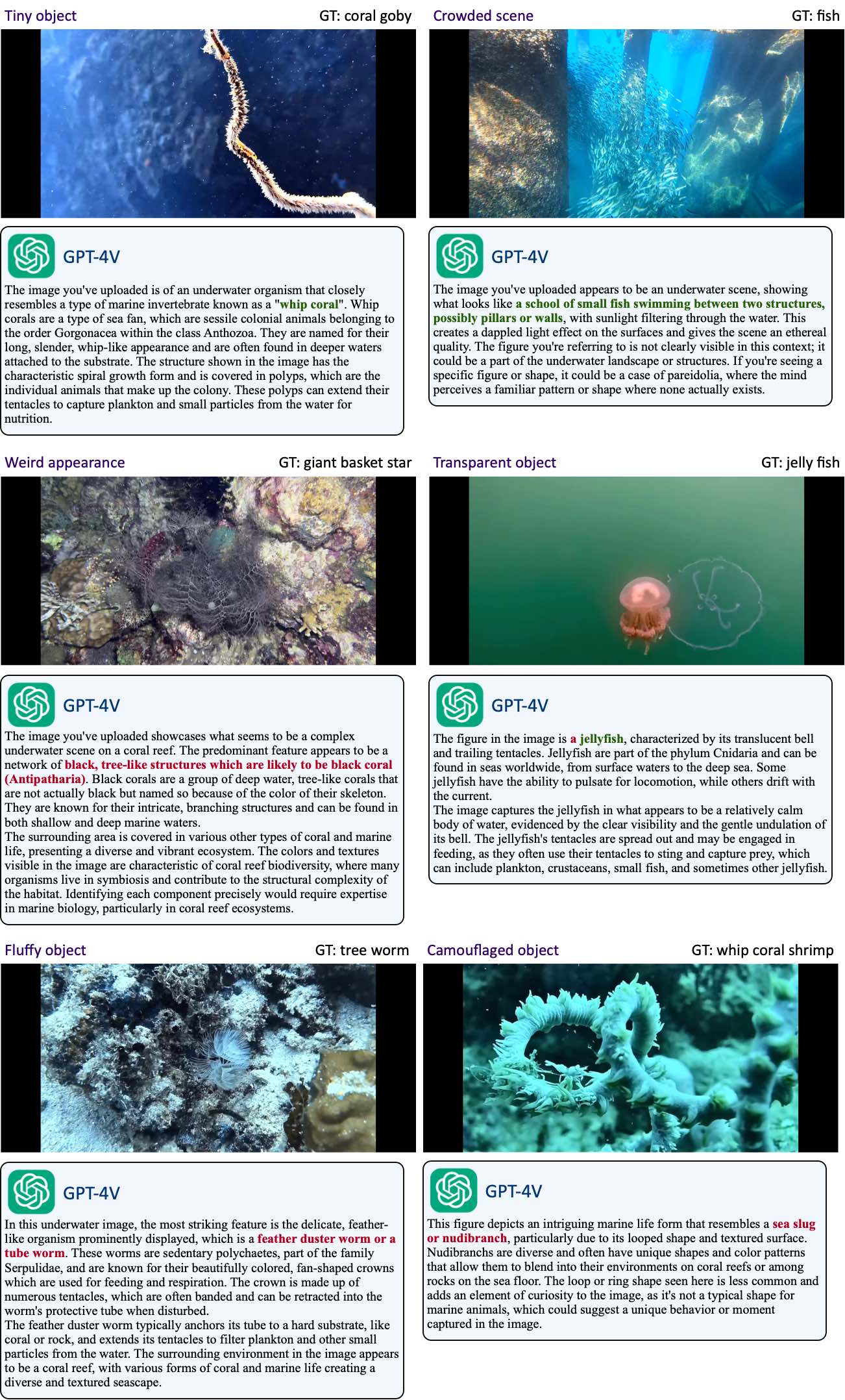}
\end{center}
\caption[Marine object recognition under challenging conditions Case 1]{The marine object recognition results under challenging conditions. The prompts are ``Recognize the object in this image and tell me the species name of the recognized objects''. The ground truths are also provided.}
\label{fig:challenge_sub1}
\end{figure}

\begin{figure}[htbp]
\vspace{-0.2in}
\begin{center}
\includegraphics[width=0.85\textwidth]{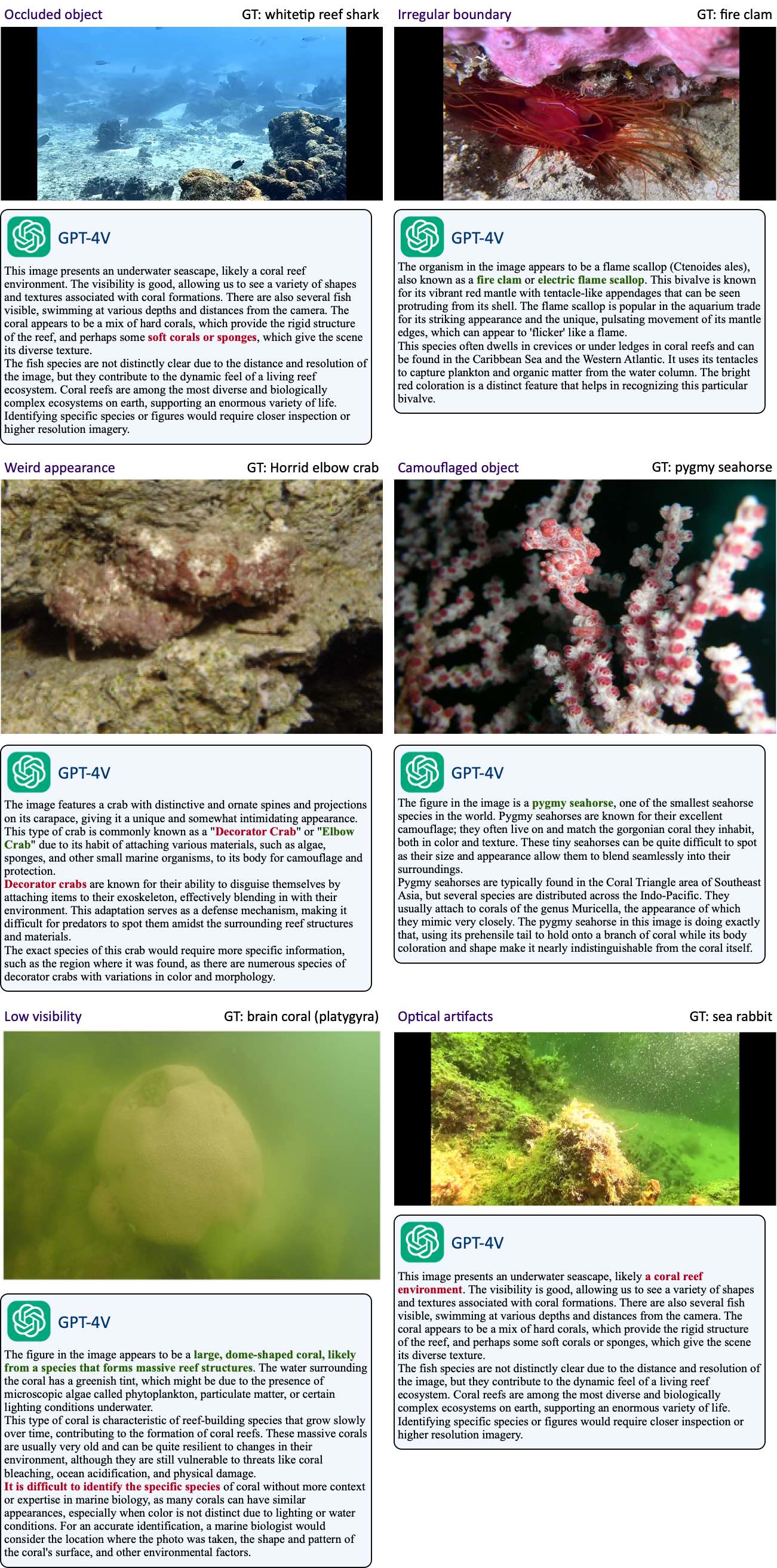}
\end{center}
\caption[Marine object recognition under challenging conditions Case 2]{The marine object recognition results under challenging conditions. The prompts are ``Recognize the object in this image and tell me the species name of the recognized objects''.}
\label{fig:challenge_sub2}
\vspace{-0.5in}
\end{figure}

\noindent\textbf{Marine object recognition under challenging conditions}. We then test whether GPT-4V is capable of depicting the key visual elements under some challenging conditions, including \textit{crowded scene}, \textit{objects with weird appearances}, \textit{fluffy object}, \textit{irregular boundary}, \textit{tiny object}, \textit{camouflaged object}, \textit{object detection under occlusion}, \textit{low visibility}, and \textit{optical artifacts}. All the experimental results are reported in Figure~\ref{fig:challenge_sub1} and Figure~\ref{fig:challenge_sub2}, respectively. For these testing experiments, we make sure there are at least 10 images under each experimental setting. We compute the recognition accuracy under those diverse settings. We observe that GPT-4V has a poor ability to accurately recognize the visual elements under challenging conditions. We guess that such failure of GPT-4V may be subject to the minority training data from the marine field. More training data collected under challenging conditions should be further included to promote the recognition ability of GPT-4V in challenging conditions. 

\newpage
\subsubsection{Fine-grained marine object recognition}
\begin{figure}[htbp]
\begin{center}
\includegraphics[width=\textwidth]{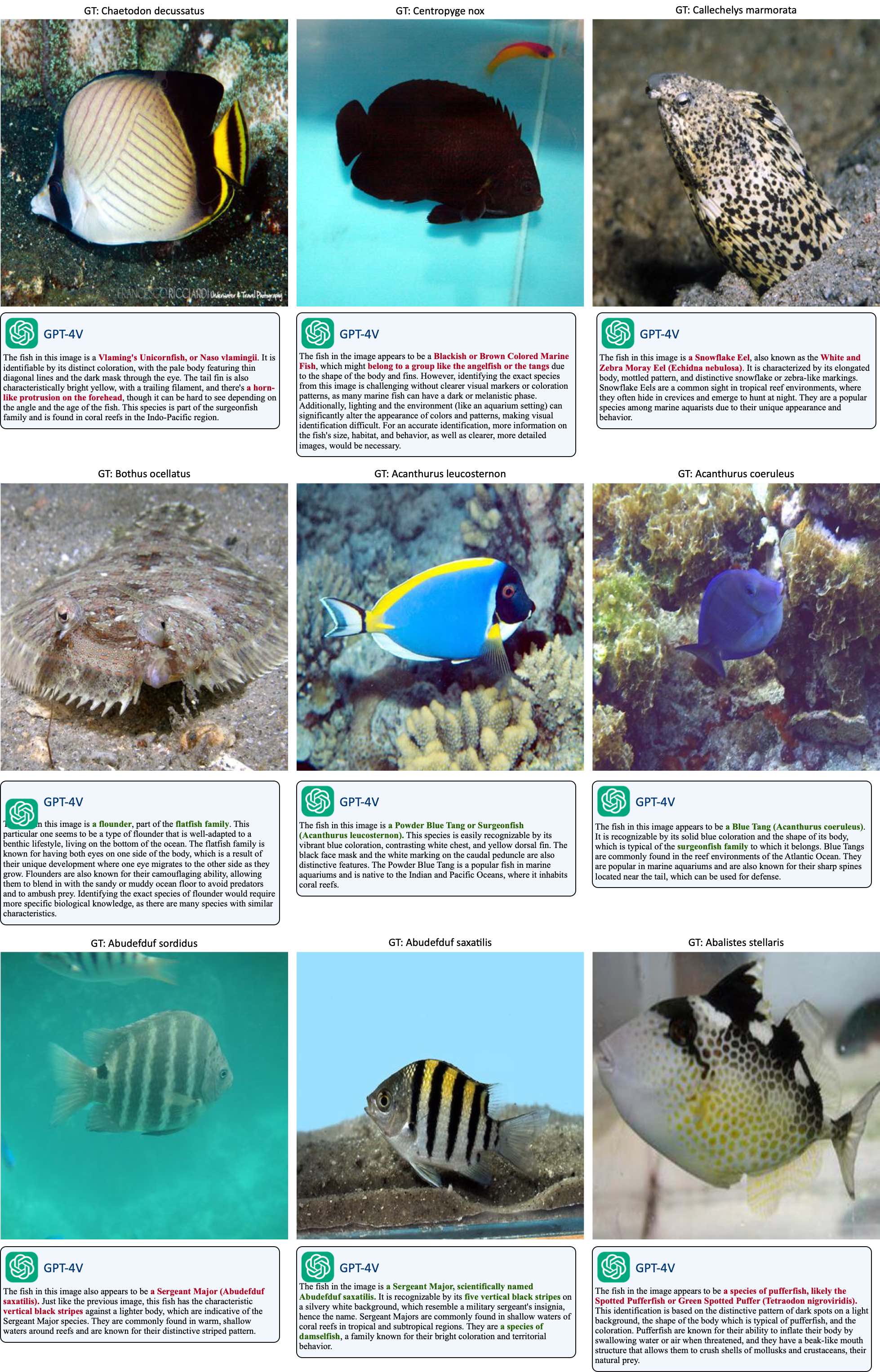}
\end{center}
\caption[Fine-grained marine object recognition]{The fine-grained marine object recognition results of GPT-4V. The prompts are ``Recognize the object in this image and tell me the species name of the recognized object''.}
\label{fig:fine}
\end{figure}

We test whether GPT-4V could discriminate very similar marine objects (\emph{e.g.}, fine-grained object recognition) and generate different responses based on given visual contents. We report the fine-grained object recognition results of GPT-4V in Figure~\ref{fig:fine}. As demonstrated, GPT-4V failed to tell the differences of close-related marine objects with similar appearances. The fine-grained object recognition ability is required in the marine analysis field since it could enable diversity monitoring and reduce the human labor from the domain experts on species identification. There is still a far away from utilizing GPT-4V for marine species identification.

\begin{figure}[htbp]
\begin{center}
\includegraphics[width=\textwidth]{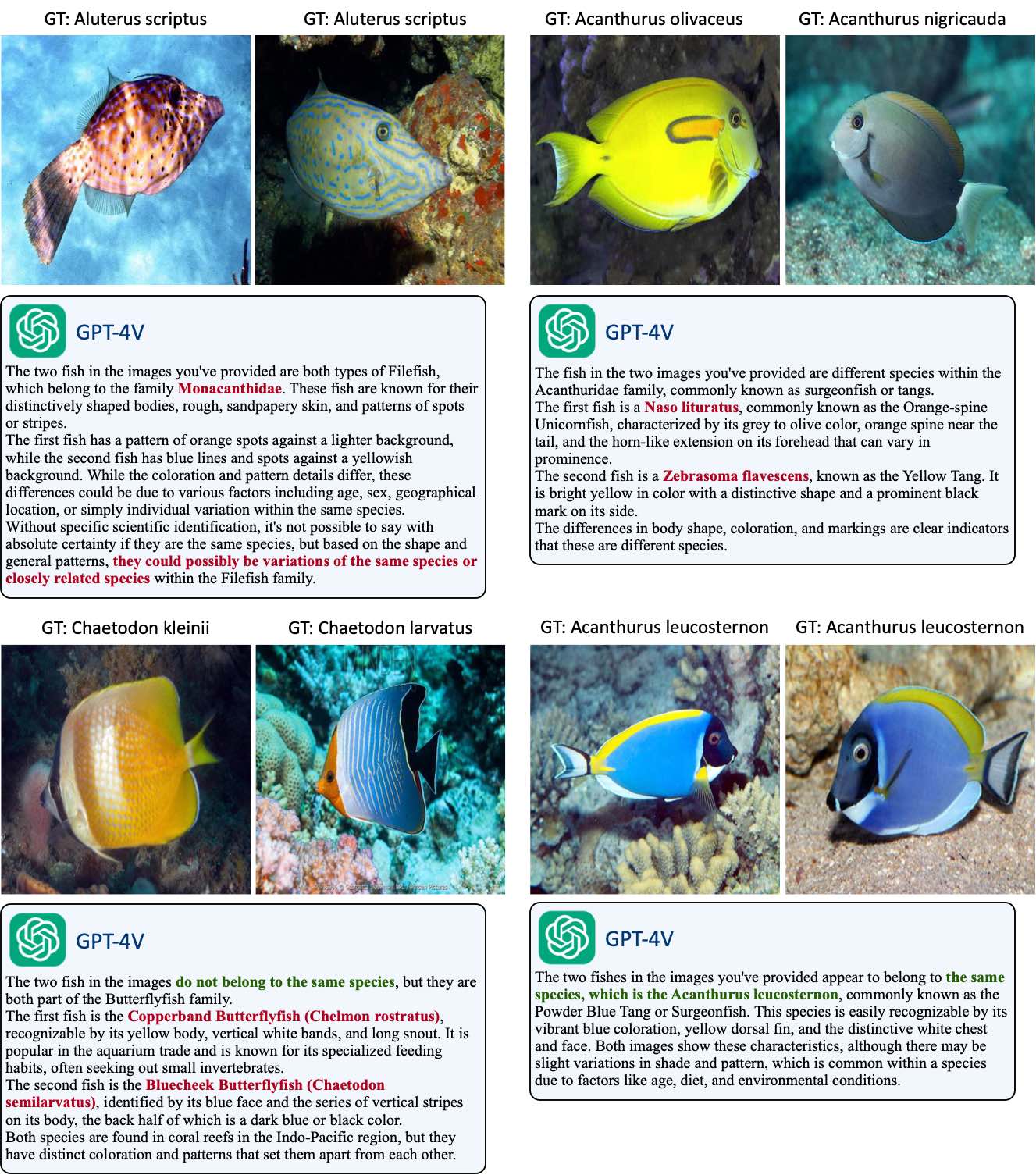}
\end{center}
\caption[Pairwise comparing]{Utilize GPT-4V for pairwise comparing. The prompts are ``Compare whether the two fishes in the two images belong to the same species''.}
\label{fig:pairwise}
\end{figure}

\begin{figure}[htbp]
\begin{center}
\includegraphics[width=\textwidth]{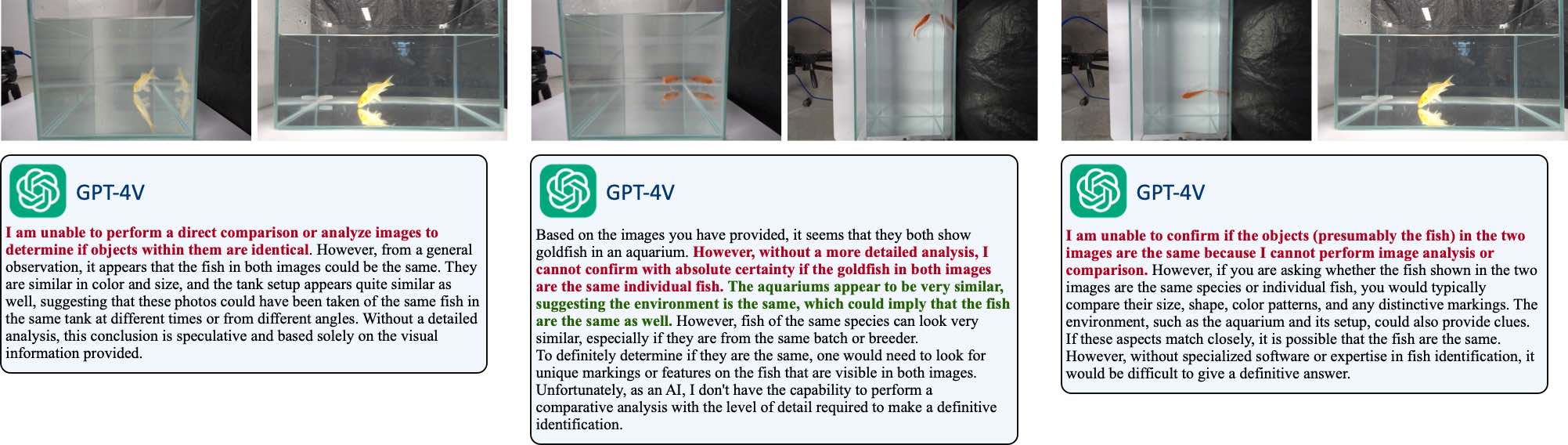}
\end{center}
\caption[Cross-view fish re-identification]{The cross-view fish re-identification performance of GPT-4V. The prompts are ``Compare whether the two fishes in the two images belong to the same species''.}
\label{fig:cross_view}
\end{figure}

We then perform the pairwise comparing, formulating a pair of images and asking GPT-4V whether the objects within the two images belong to the same marine species. Figure~\ref{fig:pairwise} illustrates the pairwise comparing performance. We formulate 20 pairs and compute the correct rate of GPT-4V on this task. \textbf{Cross-view fish re-identification}. We have also performed experiments to ask the GPT-4V to judge whether the objects within the images captured under different camera views (\emph{e.g.}, frontal, bird and side views) are the same object. Figure~\ref{fig:cross_view} demonstrates that GPT-4V has a poor ability to retrieve objects with camera view changes. GPT-4V refused to respond to the matching question even though the two fishes from the two visual images share very different appearances.

\newpage
\subsubsection{Robustness Analysis}
\begin{figure}[htbp]
\begin{center}
\includegraphics[width=\textwidth]{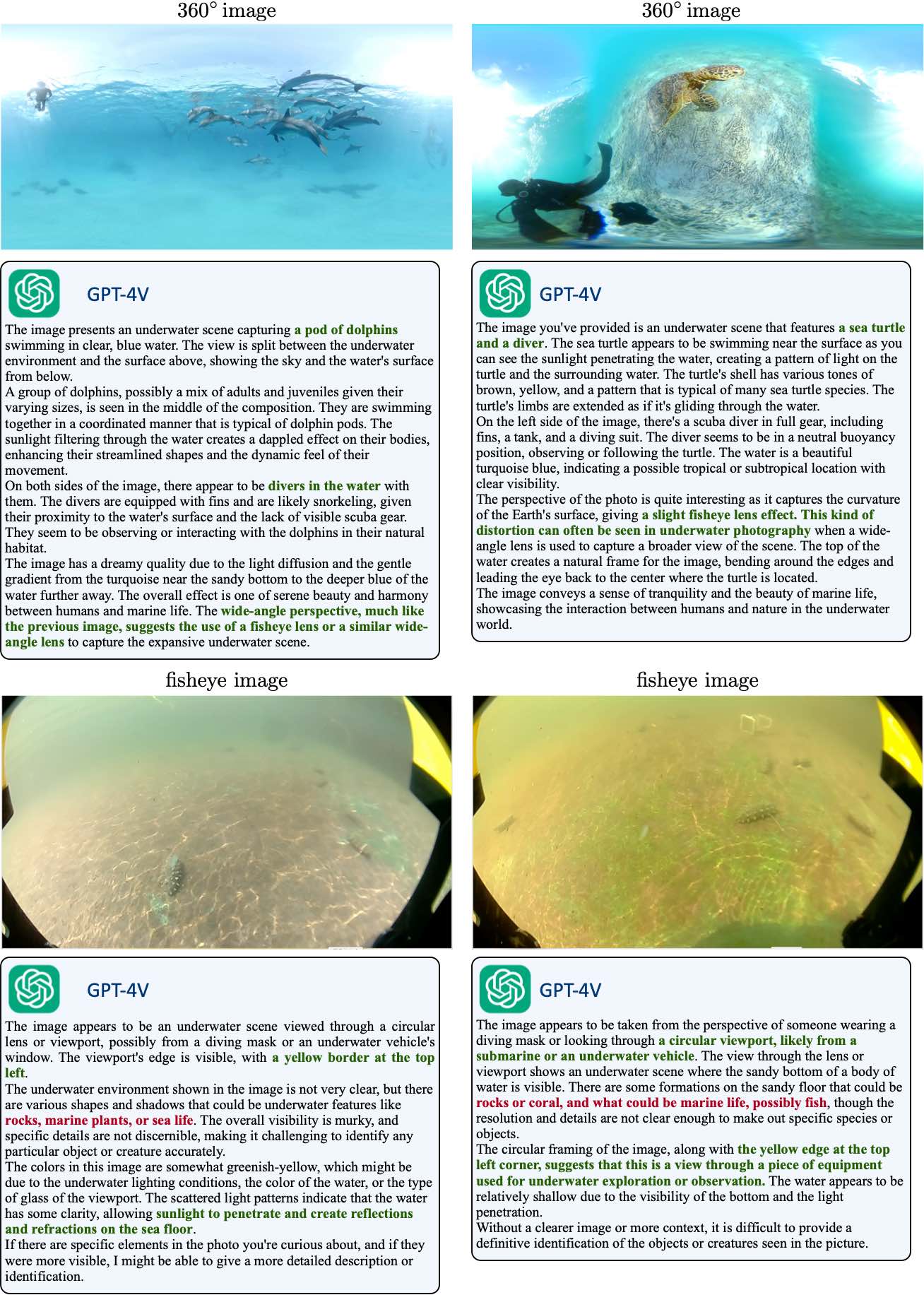}
\end{center}
\caption[Robustness analysis Case 1]{The marine object recognition of GPT-4V on 360$^{\circ}$ and fisheye images. The prompts are ``Describe this figure in detail and recognize the object within this figure''.}
\label{fig:robust_sub1}
\end{figure}

\begin{figure}[htbp]
\begin{center}
\includegraphics[width=0.9\textwidth]{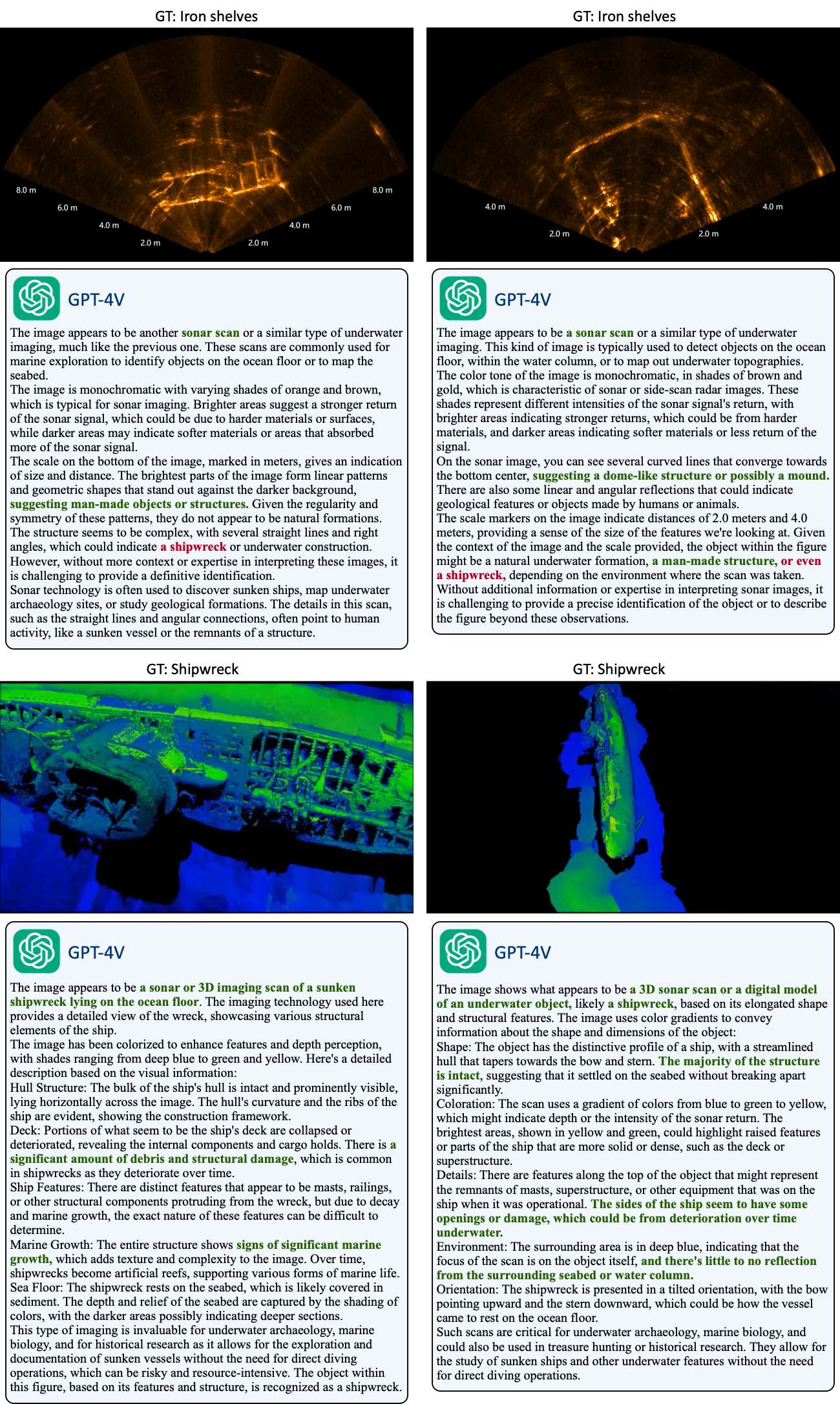}
\end{center}
\caption[Robustness analysis Case 2]{The marine object recognition of GPT-4V on sonar and lidar images. The prompts are ``Describe this figure in detail and recognize the object within this figure''.}
\label{fig:robust_sub2}
\end{figure}

In this section, we test the robustness of GPT-4V in recognizing various formats of visual signals, such as the fisheye~\citep{zheng2023real}, 360$^{\circ}$~\citep{huang2023360vot}, sonar~\citep{xie2022dataset} and Lidar images. Figure~\ref{fig:robust_sub1} illustrates the recognition results of GPT-4V on 360$^{\circ}$ and fisheye images. GPT-4V could observe the distortion of 360$^{\circ}$ images but cannot explicitly explain why the distortion happens. In most cases, it could accurately recognize the visual elements from the visual images, however, it seems to have hallucination on the components in the submarine images where the visibility is low and images tend to be more murky, showing its limited robustness to fisheye and 360$^{\circ}$ images. What's more, it is an expert at recognizing how the images are captured through the edge or border of the viewpoint. We report the further object recognition results of GPT-4V on sonar images and Lidar images in Figure~\ref{fig:robust_sub2}. GPT-4V can recognize the general shape of the existing objects but cannot effectively detect what kind of stuff they are in sonar images due to the appearance shift. But for Lidar images in which objects' appearance doesn't shift a lot, GPT-4T can exactly describe the element in detail, showing a very good understanding of the image.

\begin{figure}[htbp]
\begin{center}
\includegraphics[width=\textwidth]{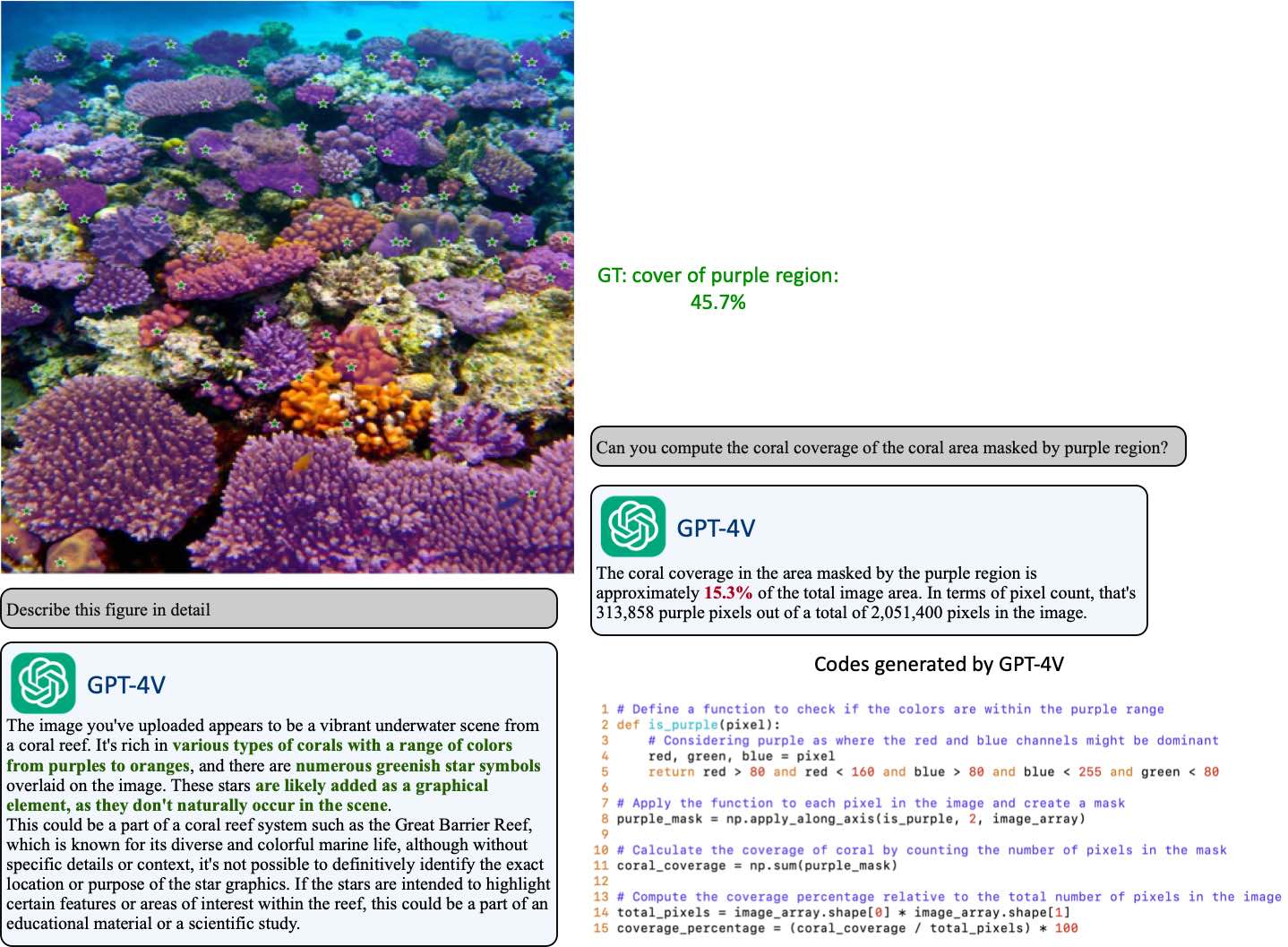}
\end{center}
\caption[Robustness analysis Case 3]{The marine object recognition results of GPT-4V on the images with highlighted regions.}
\label{fig:highlight}
\end{figure}

We then identify whether GPT-4V could effectively recognize object regions with highlighted masks as demonstrated in Figure~\ref{fig:highlight}, exploring the referring comprehension ability of GPT-4V. The partial parts of the whole image are highlighted by purple and we ask GPT-4V to identify the highlighted regions. Furthermore, GPT-4V is asked to compute the cover of the highlighted coral regions. GPT-4V could generate the Python codes to compute the cover statistics. However, GPT-4V would self-define the RGB value range of ``purple'' without explanation. However, such a definition could be wrong and cannot handle visual images with high complexity.

\newpage
\subsubsection{Physical World Knowledge Understanding}
\begin{figure}[htbp]
\begin{center}
\includegraphics[width=\textwidth]{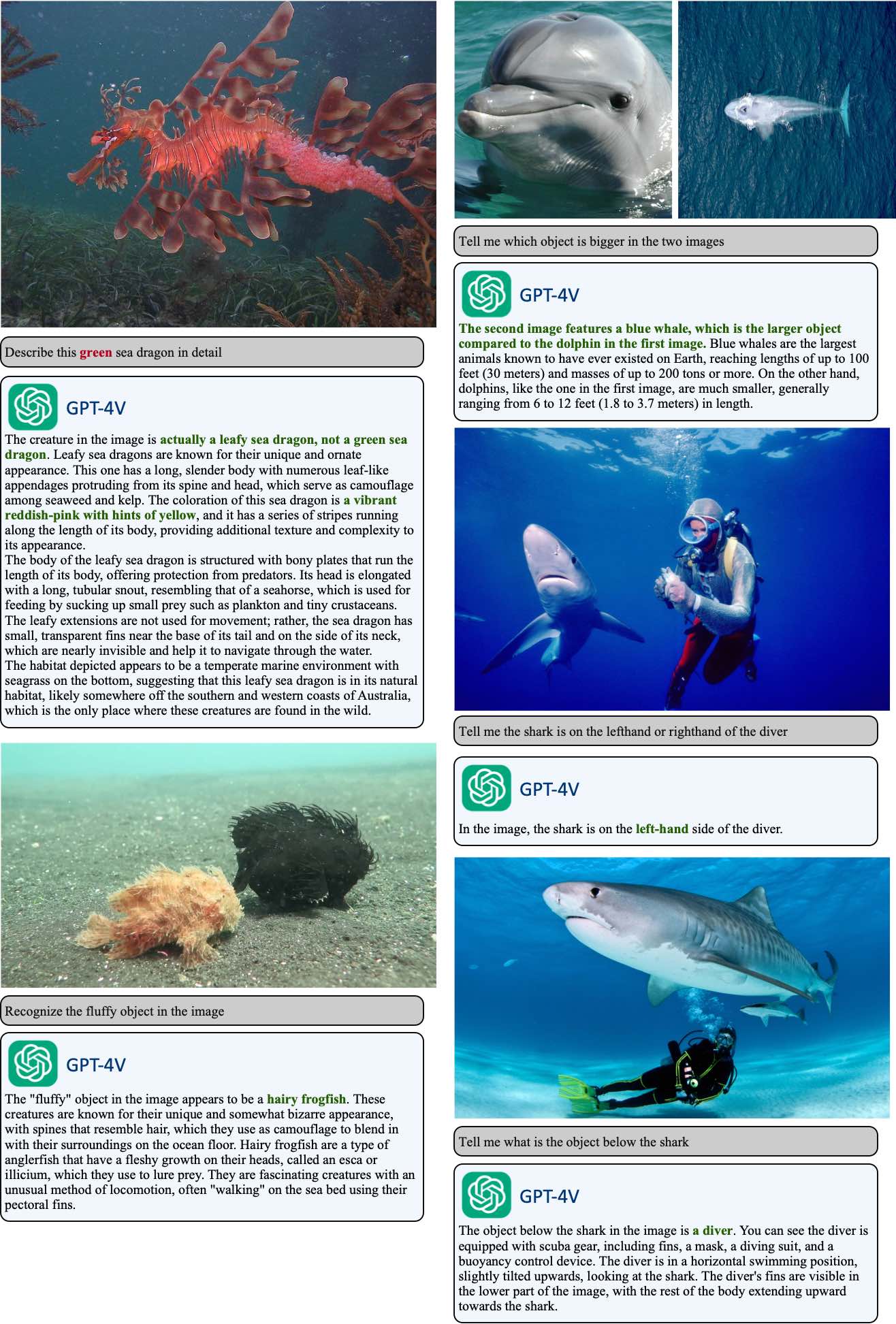}
\end{center}
\caption[Physical world knowledge understanding]{GPT-4V could understand the physical world knowledge.}
\label{fig:physical}
\end{figure}

We finally explore whether GPT-4V could really understand the physical world knowledge, for example, the spatial, size, color and texture attributes of the existing objects within the images. We explore the capability of GPT-4V to apply common sense knowledge in understanding visual contents within images. We have investigated the models' ability to comprehend visual information via the application of knowledge, which encompasses commonsense, subject knowledge, multicultural customs, and world knowledge. The results are illustrated in Figure~\ref{fig:physical}. GPT-4V shows its strong capability of understanding the physical world knowledge like spatial, size and texture attributes and it also has great robustness to the wrong knowledge that does not correspond with the image and correct it. Even if we provide it with some really misleading images with close view of a dolphin and a far view of a blue whale, it could still correctly tell the real size of these objects.

\newpage
\subsection{Statistics}
In this section, we aim to explore the ability of GPT-4V to perform visual statistics based on the visual contents, such as object counting and summarizing all the appeared objects within images. 

\subsubsection{Object counting}
\begin{figure}[htbp]
\vspace{-0.2in}
\begin{center}
\includegraphics[width=0.88\textwidth]{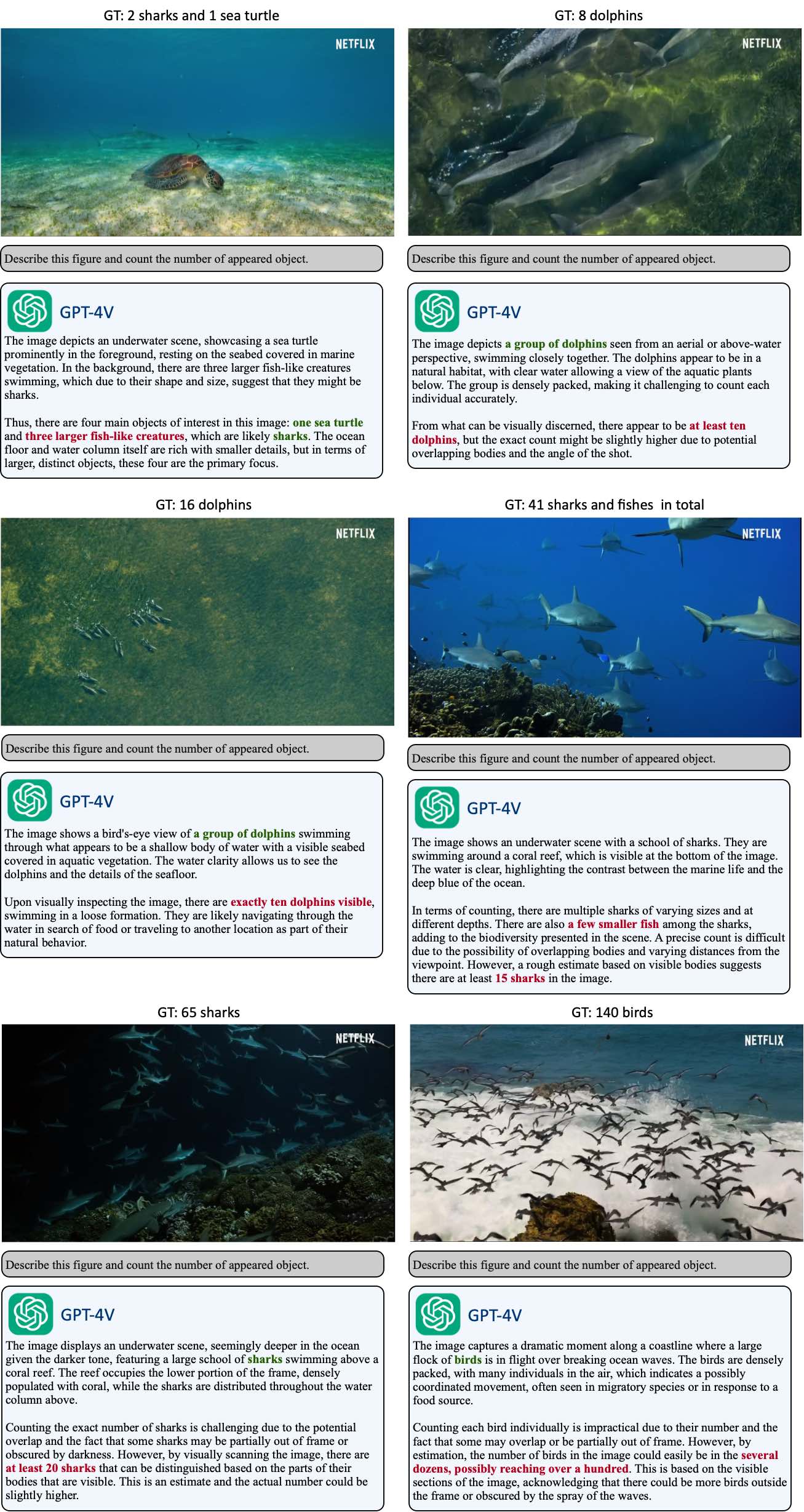}
\end{center}
\caption[Object counting]{The marine object counting results under different settings. }
\label{fig:counting}
\vspace{-0.2in}
\end{figure}

We perform object counting experiments under five settings: 1) fewer than 10 objects; 2) 10-20 objects; 3) 20-50 objects; 4) 50-100 objects and 5) more than 100 objects. All the qualitative results have been reported in Figure~\ref{fig:counting}. As demonstrated, GPT-4V only demonstrates a limited ability to count the existing objects within the images, especially if the objects are occluded together or the objects are tiny. Meanwhile, since the GPT-4V directly yields the estimation results of objects without explicitly localizing the objects (\emph{e.g.}, bounding box), the estimation results will likely be not accurate. Furthermore, we have also observed that GPT-4V tends to generate an exact number of presented objects within the images when there are few objects visible. In contrast, GPT-4V instead yields a rough number of the object counting results. To avoid potential mistakes, GPT-4V outputs a range (\emph{e.g.}, more than 100) for the estimated objects. In summary, the external object detection tools for localizing the objects should be integrated to promote the object counting ability of MLLM.

\subsubsection{Recognizing all the objects}
\begin{figure}[htbp]
\begin{center}
\includegraphics[width=\textwidth]{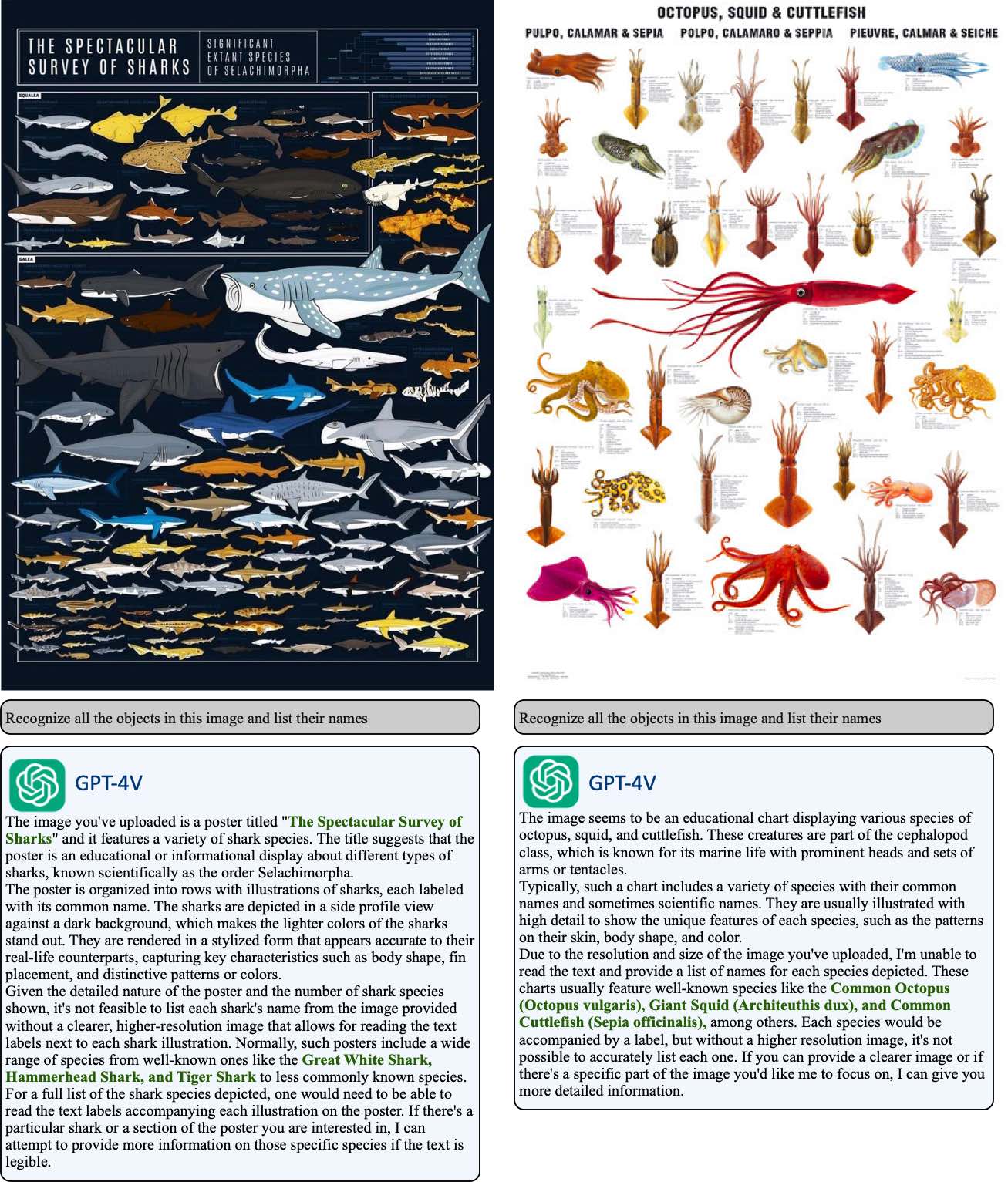}
\end{center}
\caption[Recognizing all the objects within images]{Utilize GPT-4V to recognize all the objects within the visual images.}
\label{fig:recognize_all}
\end{figure}

We then explore the ability of GPT-4V to recognize all the existing objects within the given visual images and list the corresponding names of all the recognized objects. Figure~\ref{fig:recognize_all} demonstrates the recognition results under the crowded and structured palette. The GPT-4V struggles to recognize all the objects within the images and only lists very few common object categories. Furthermore, we observe that GPT-4V could summarize the implicit intention of such visual images and try to summarize the relationships between the recognized objects. However, due to the large number of objects, some less commonly known species, and the low image resolution, GPT-4V shows a very limited performance on recognizing all the objects in one single image while it could still understand some general information of the image, like title, colors and common features of objects. Similar to the object counting task, GPT-4V tends to discard many objects within the images and only tries to recognize some common objects easy to recognize to avoid making mistakes, but this also makes it hard for GPT-4V to recognize all the objects existing in the image.

\newpage
\subsection{Domain-specific Question-Answering}
we examine the ability of GPT-4V to apply knowledge in the fields of marine to understand visual images. We observe that GPT-4V possesses the relevant subject knowledge associated with the following cases.
\begin{figure}[htbp]
\begin{center}
\includegraphics[width=\textwidth]{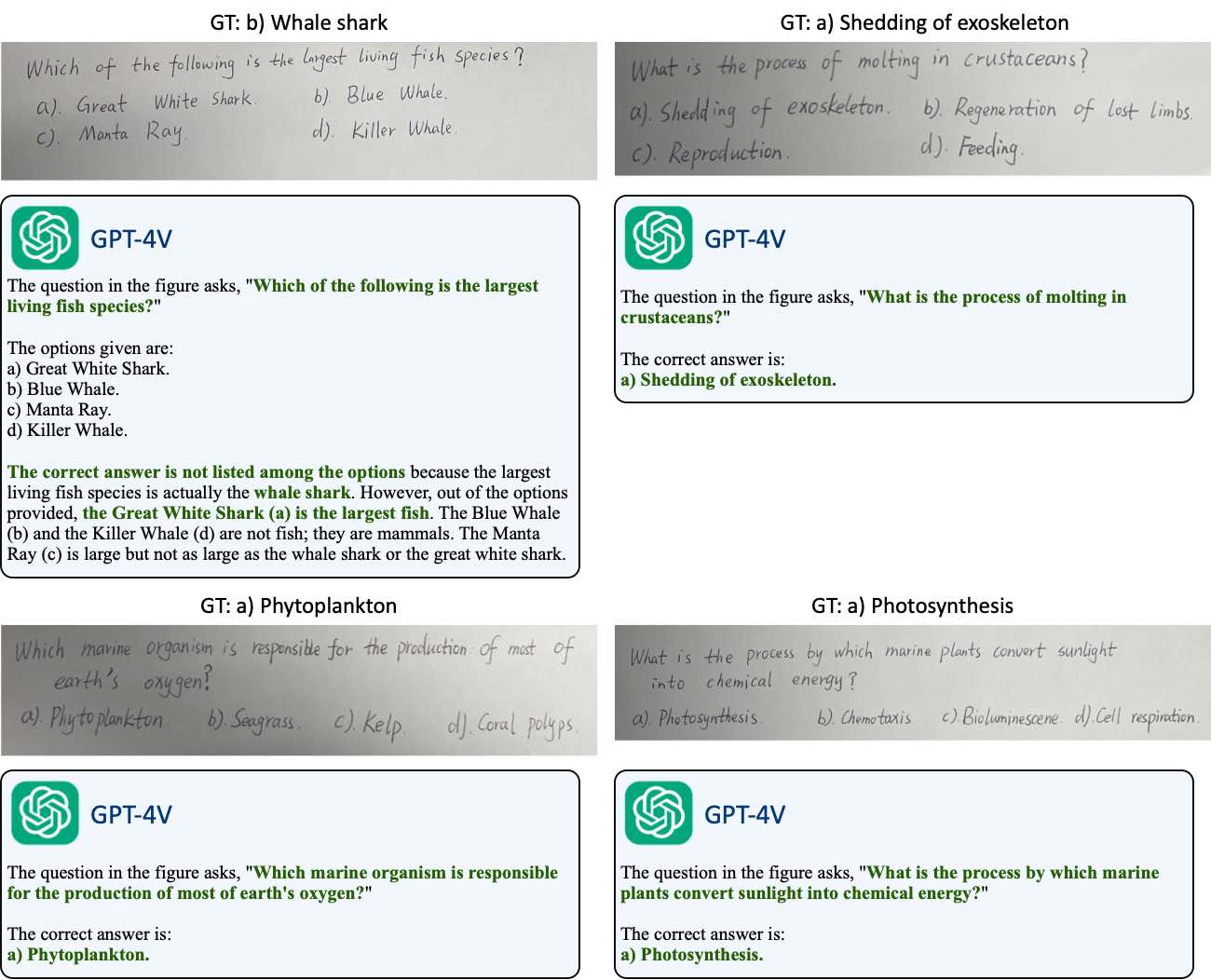}
\end{center}
\caption[Marine multiple choice question answering]{The performance of GPT-4V on answering the marine multiple choice questions. The prompts are ``answer the question within this image''. We observe that GPT-4V demonstrates a strong OCR ability.}
\label{fig:ocr_mcq}
\end{figure}

\noindent\textbf{Multiple choice questions}. We first explore the ability of GPT-4V to answer the marine multiple-choice questions. We upload the manually written marine questions and corresponding choices to GPT-4V and ask GPT-4V to generate the answers in Figure~\ref{fig:ocr_mcq}. As demonstrated, GPT-4V has shown a strong optical character recognition (OCR) ability to extract the correct text information from the uploaded images and a satisfactory promise for handling basic marine knowledge. We have manually constructed 100 multiple-choice questions, which come from marine biology, oceanography, and geology. The accuracy of GPT-4V is computed to quantitatively assess the quality of GPT-4V in answering the domain-specific questions. 

\begin{figure}[htbp]
\begin{center}
\includegraphics[width=\textwidth]{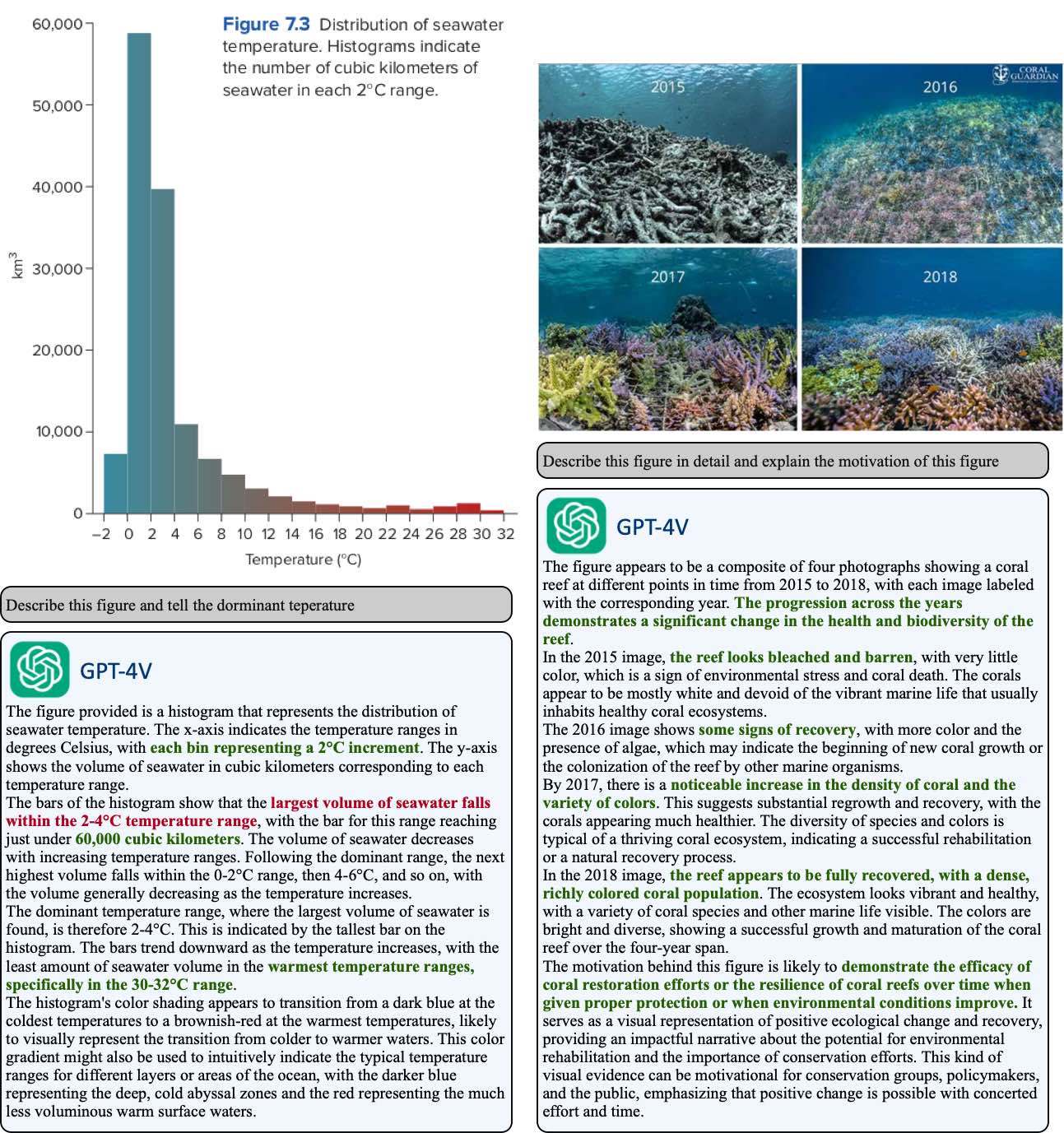}
\end{center}
\caption[Domain-specific question answering]{The performance of GPT-4V on answering domain-specific questions.}
\label{fig:vqa_sub1}
\end{figure}

\begin{figure}[htbp]
\begin{center}
\includegraphics[width=\textwidth]{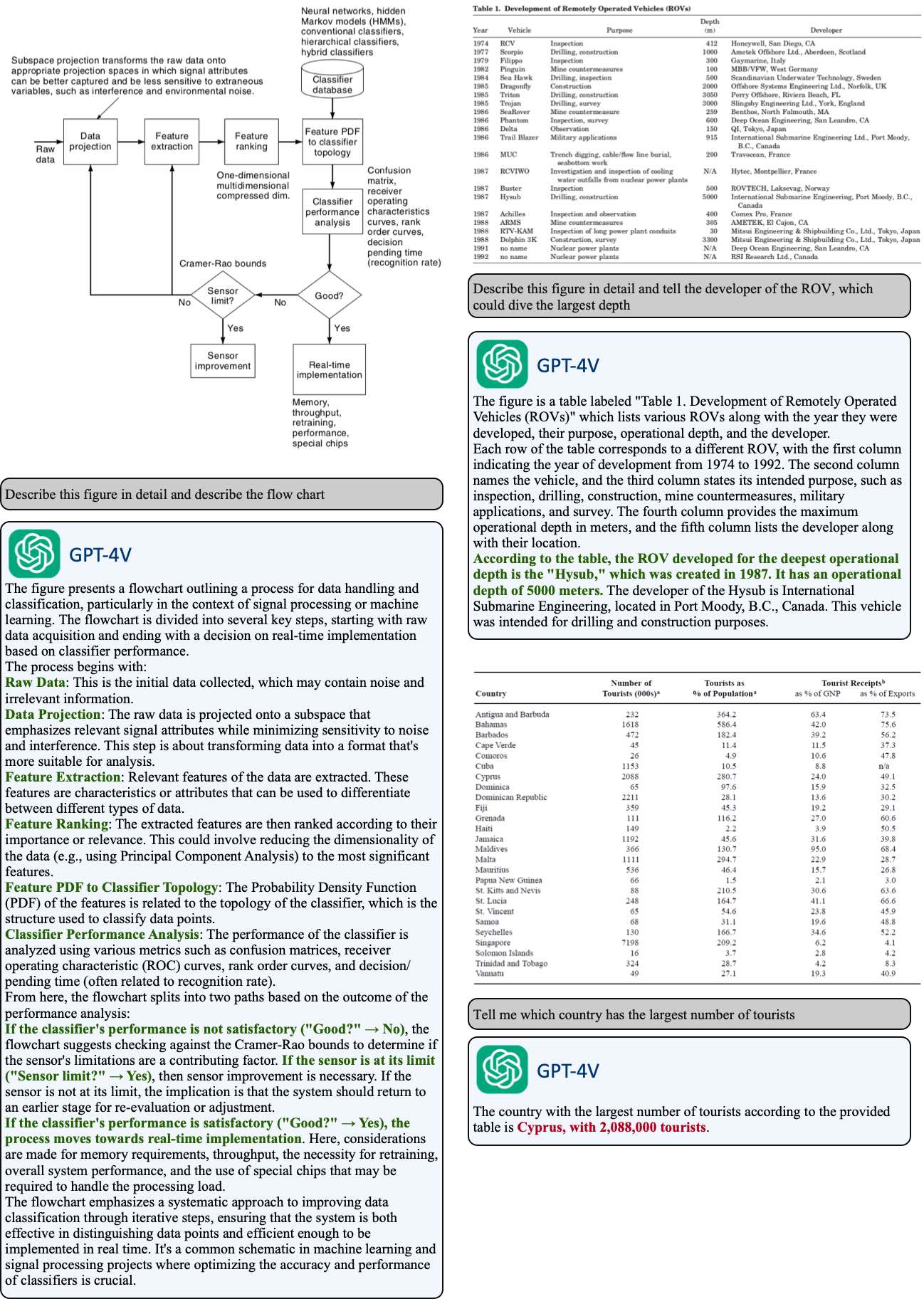}
\end{center}
\caption[Understanding domain-specific figures and tables]{The performance of GPT-4V on understanding domain-specific figures and tables.}
\label{fig:vqa_sub2}
\end{figure}

\noindent\textbf{Domain-specific VQA}. We evaluate whether GPT-4V could understand the user intent of the domain experts and the ability of GPT-4V for abstract visual reasoning and scientific problem-solving. Such abilities are required for marine researchers to analyze the data (figures and tables) collected to gain insights into various aspects of marine research fields. Results are reported in Figure~\ref{fig:vqa_sub1} and Figure~\ref{fig:vqa_sub2}, respectively. As demonstrated in Figure~\ref{fig:vqa_sub1}, GPT-4V could understand most elements of the left scientific figure but make a tiny mistake about the temperature range. Besides, GPT-4V could understand the temporal changes within the scientific figure and conclude the implicit intention. It could accurately describe the coral status of each sub-figure and conclude the progression changes. We have also included more visual scientific examples essential for handling marine biology, engineering, oceanography, and \textit{etc}. 

Furthermore, we feed GPT-4V with scientific figures and tables from the field of marine engineering as reported in Figure~\ref{fig:vqa_sub2}. GPT-4V could effectively understand the flowchart. GPT-4V could describe the logic inside of the flowchart and respond with more reasoning details. GPT-4V could also understand the tables in detail. When being asked a question that requires intermediate reasoning procedures, GPT-4V could answer correctly with detailed reasoning procedures. However, GPT-4V still has difficulties in providing a precise answer in some cases, which is mainly constrained by the unsatisfactory OCR accuracy in Figure~\ref{fig:vqa_sub2}.

\begin{figure}[htbp]
\begin{center}
\includegraphics[width=\textwidth]{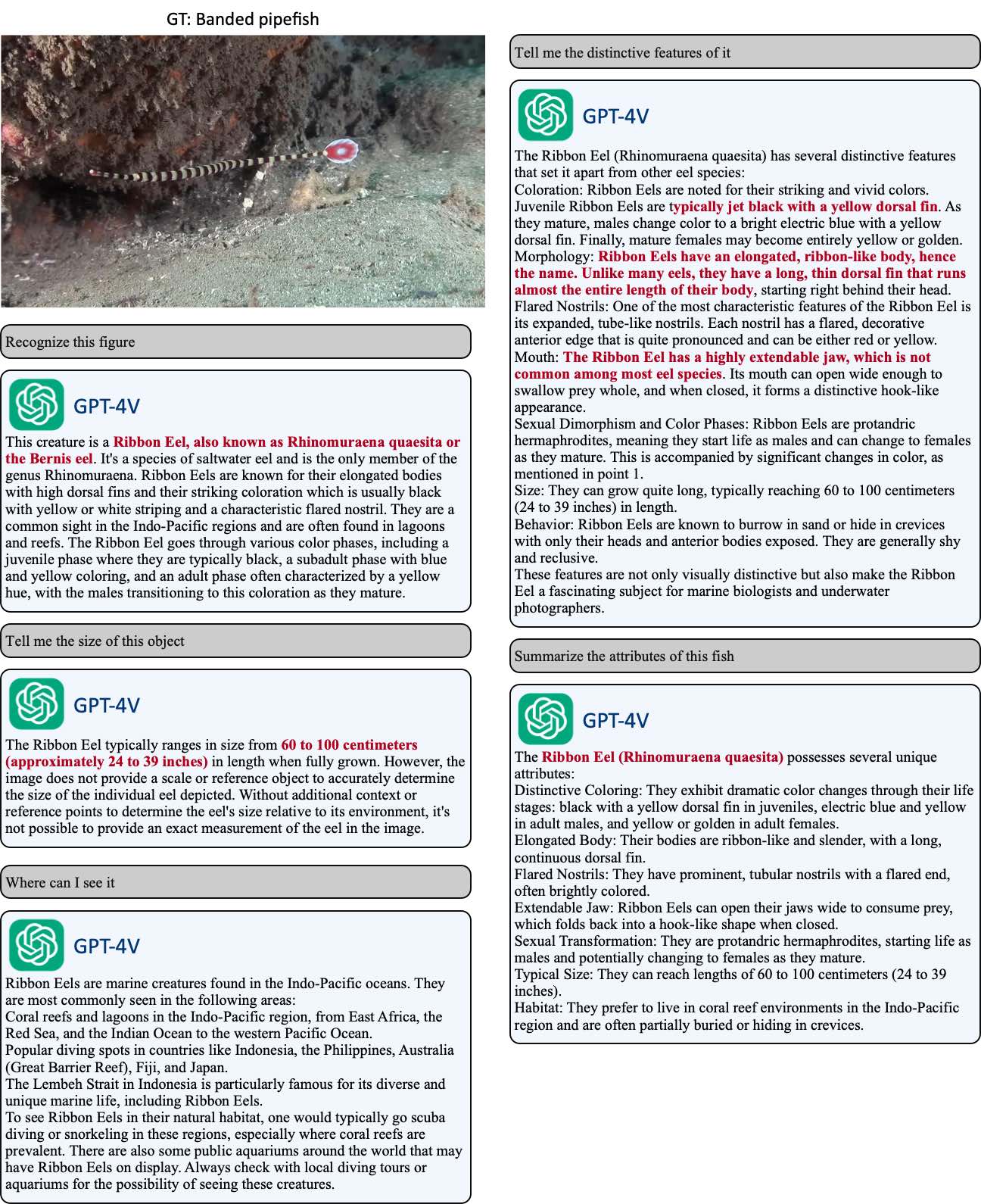}
\end{center}
\caption[Multi-round conversation]{The GPT-4V could support multi-round conversation, however, leads to error accumulation.}
\label{fig:multi}
\end{figure}

\noindent\textbf{Multi-round conversation}. We finally assess the ability of GPT-4V to support multi-round conversations. Users could ask different questions for comprehensive analysis, as demonstrated in Figure~\ref{fig:multi}. Our study suggests that GPT-4V, could generate corresponding responses aligned with the user intent and cover the detailed information. However, GPT-4V struggles with the marine object recognition. With the wrongly identified marine objects, GPT-4V leads to error accumulation, which suggests that GPT-4V only responds based on the previously generated keywords (as the context prompt) without looking at the visual contents. How to alleviate the hallucination of MLLMs is a valuable and important future research direction.

\newpage
\subsection{Marine Cultural Understanding}
\begin{figure}[htbp]
\begin{center}
\includegraphics[width=\textwidth]{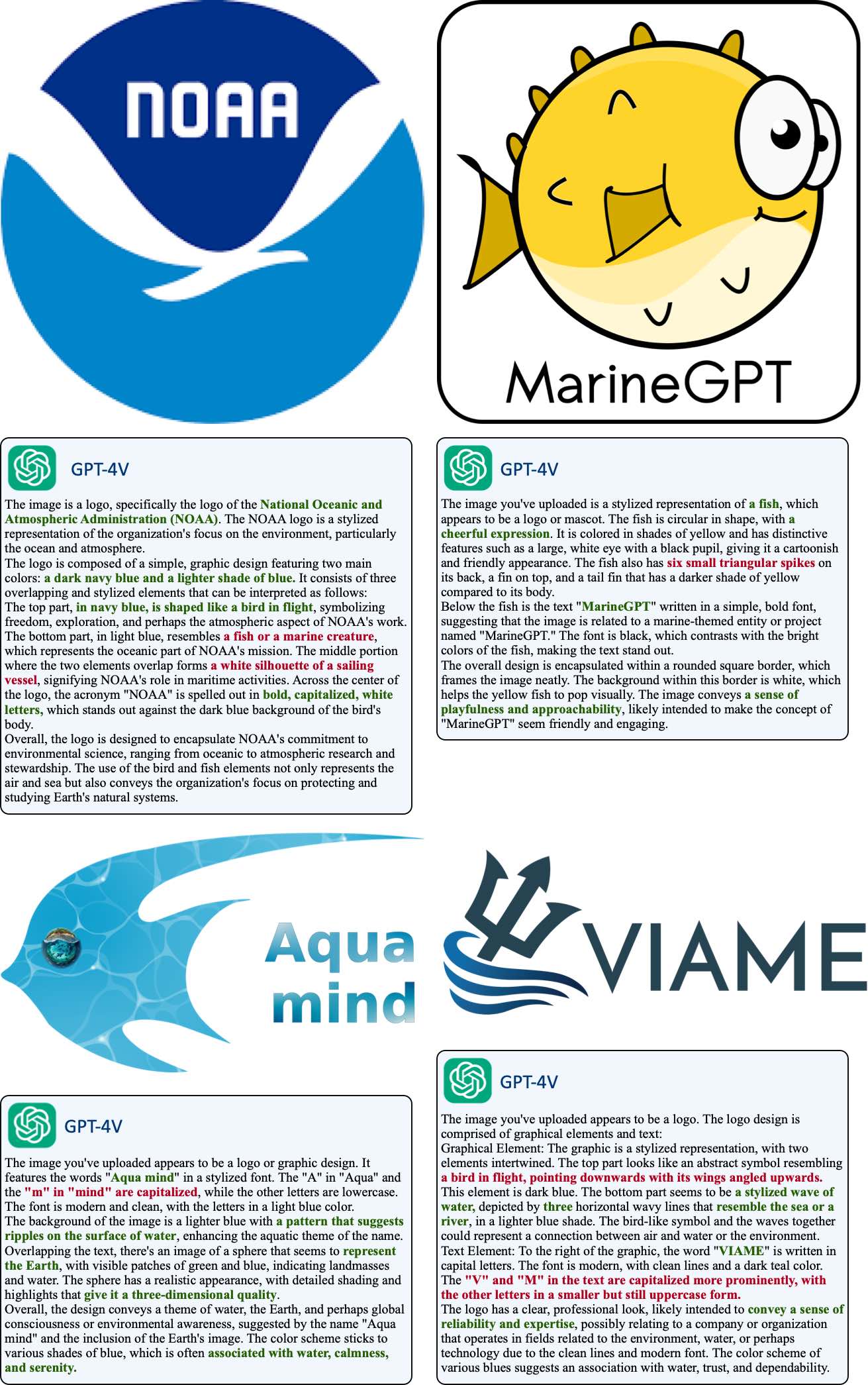}
\end{center}
\caption[Marine logo understanding]{Utilize GPT-4V for marine logo understanding.}
\label{fig:marine_logo}
\end{figure}

\begin{figure}[htbp]
\begin{center}
\includegraphics[width=0.9\textwidth]{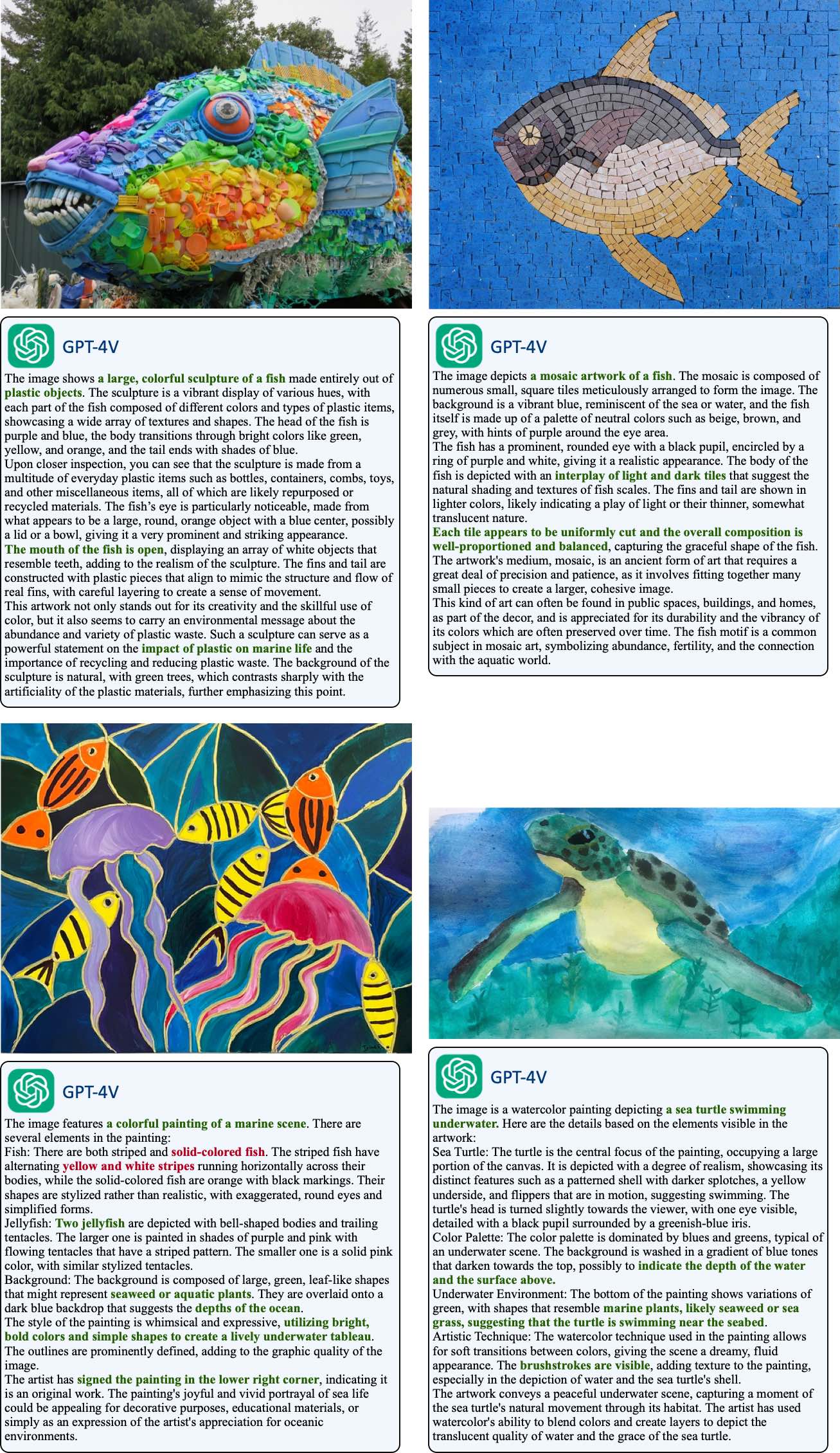}
\end{center}
\caption[Artist image understanding]{Utilize GPT-4V for marine artist image understanding.}
\label{fig:marine_art}
\end{figure}

We investigate the ability of GPT-4V to recognize logos, landmarks, artist images, and more in Figure~\ref{fig:marine_logo}, Figure~\ref{fig:marine_art}, and Figure~\ref{fig:marine_art}. 

In Figure~\ref{fig:marine_logo}, GPT-4V could effectively recognize the globally known NOAA logo and yield a detailed description of the appearance of the logo. However, there is still a hallucination with the description of the NOAA logo. We guess the generated responses are from the training corpus of GPT-4V rather than being aligned with the visual elements. As for the novel logos, GPT-4V could describe the appearance of the designed logos. The feature patterns of the logos are comprehensively described and GPT-4V could assess the artistic and literary representations of themes and species.

We then ask GPT-4V to perform marine artist image recognition and description as illustrated in Figure~\ref{fig:marine_art}. GPT-4V could efficiently describe the visual elements of marine artist images. We present the capacity of GPT-4V to depict the appearance of the cartoon images, paintings, and actual photographs. GPT-4V demonstrates a strong ability to assess the aesthetic quality of visual images and describe the partial parts of each image. 

\begin{figure}[htbp]
\begin{center}
\includegraphics[width=\textwidth]{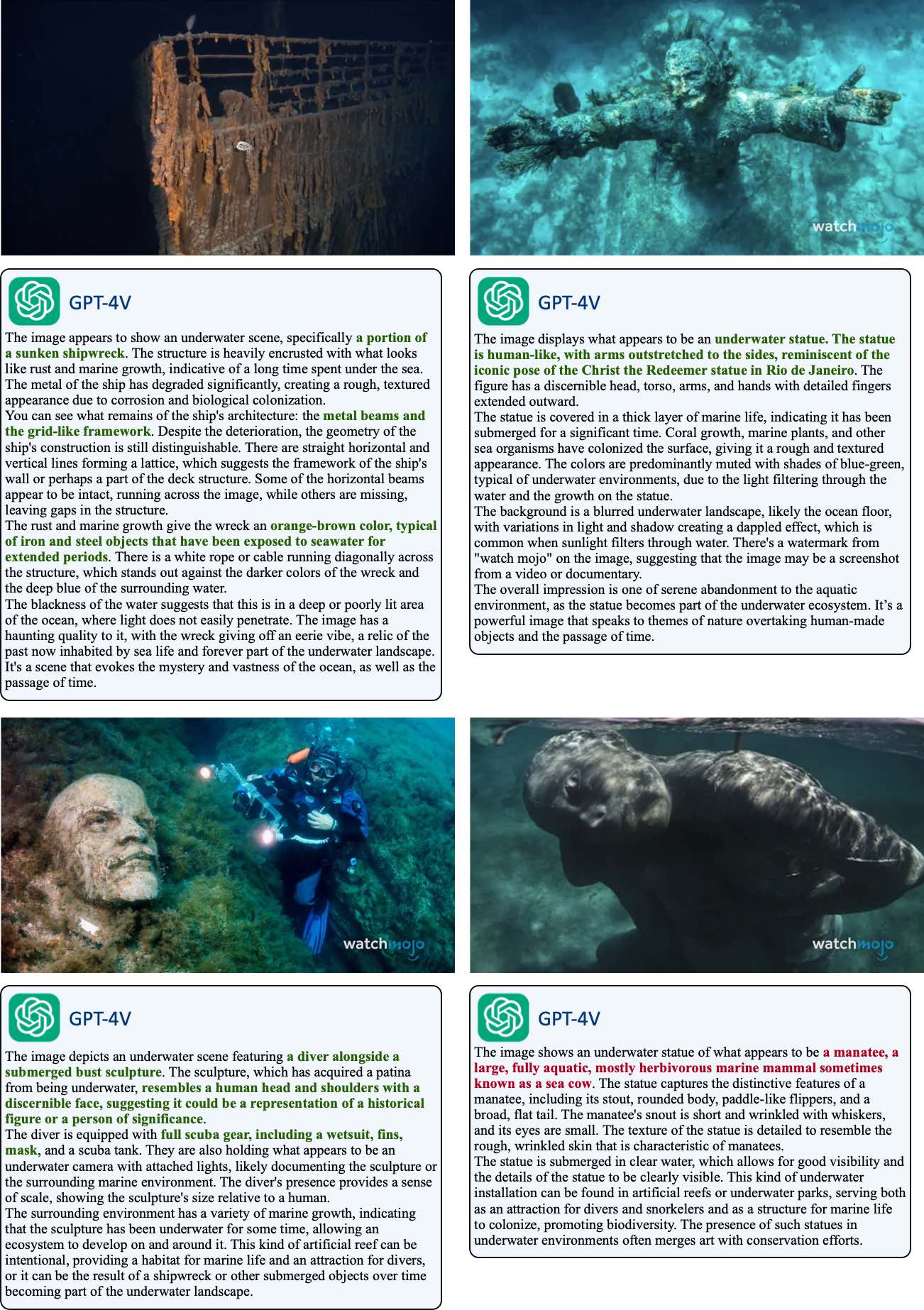}
\end{center}
\caption[Landmark recognition]{Utilize GPT-4V for marine landmark recognition.}
\label{fig:marine_landmark}
\end{figure}
Finally, we report the landmark recognition performance of GPT-4V in Figure~\ref{fig:marine_landmark}. GPT-4V can identify the marine vestige and statures. The detailed appearances of recognized ruins are further described in detail, demonstrating the strong ability of GPT-4V to perceive the visual images. However, GPT-4V cannot accurately discriminate the statures with irregular shapes and poses. 

\newpage
\subsection{Advanced Functions}
In this section, we aim to explore the possibility of utilizing GPT-4V for some advanced and complicated functions in the marine research field, such as coral coverage estimation, benthic composition statistic, multi-modal reasoning, relationship summarization, and \textit{etc}.

\subsubsection{Coral coverage estimation}
\begin{figure}[htbp]
\begin{center}
\includegraphics[width=\textwidth]{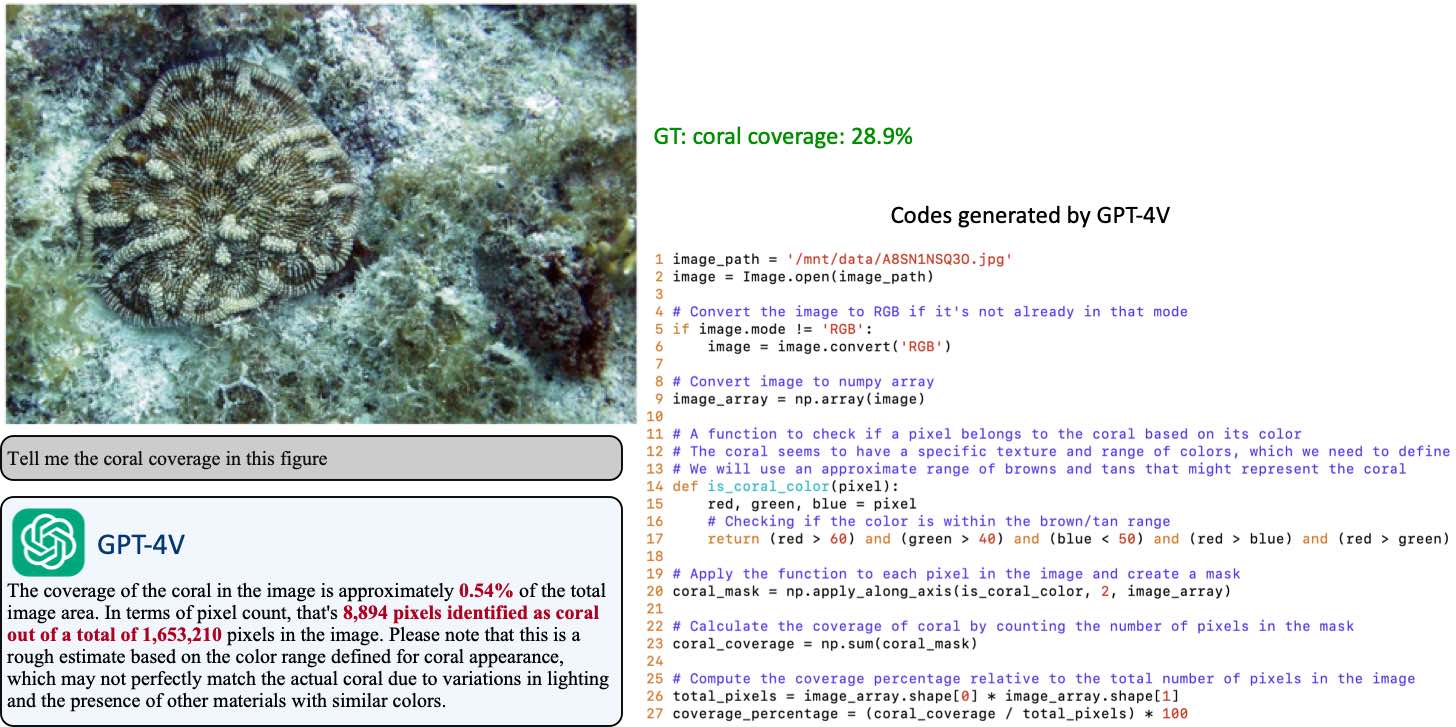}
\end{center}
\caption[Coral coverage estimation]{Utilize GPT-4V for coral coverage estimation.}
\label{fig:coverage}
\end{figure}

Coral reefs are among the most biodiverse ecosystems on our planet and provide habitat for countless marine species. Monitoring coral coverage allows researchers to assess the overall health and condition of these ecosystems. In this section, we aim to explore the feasibility of utilizing GPT-4V for coral coverage estimation. Figure~\ref{fig:coverage} represents some preliminary results of coral coverage estimation. GPT-4V avoids directly outputting the coral coverage and instead attempts to generate some computer vision processing codes for coral coverage estimation. The generated coral coverage is far away from the real ground truth. Besides, GPT-4V may lead to the ignorance of the tiny corals or the minority coral types and then result in wrong policy making.

\begin{figure}[htbp]
\begin{center}
\includegraphics[width=\textwidth]{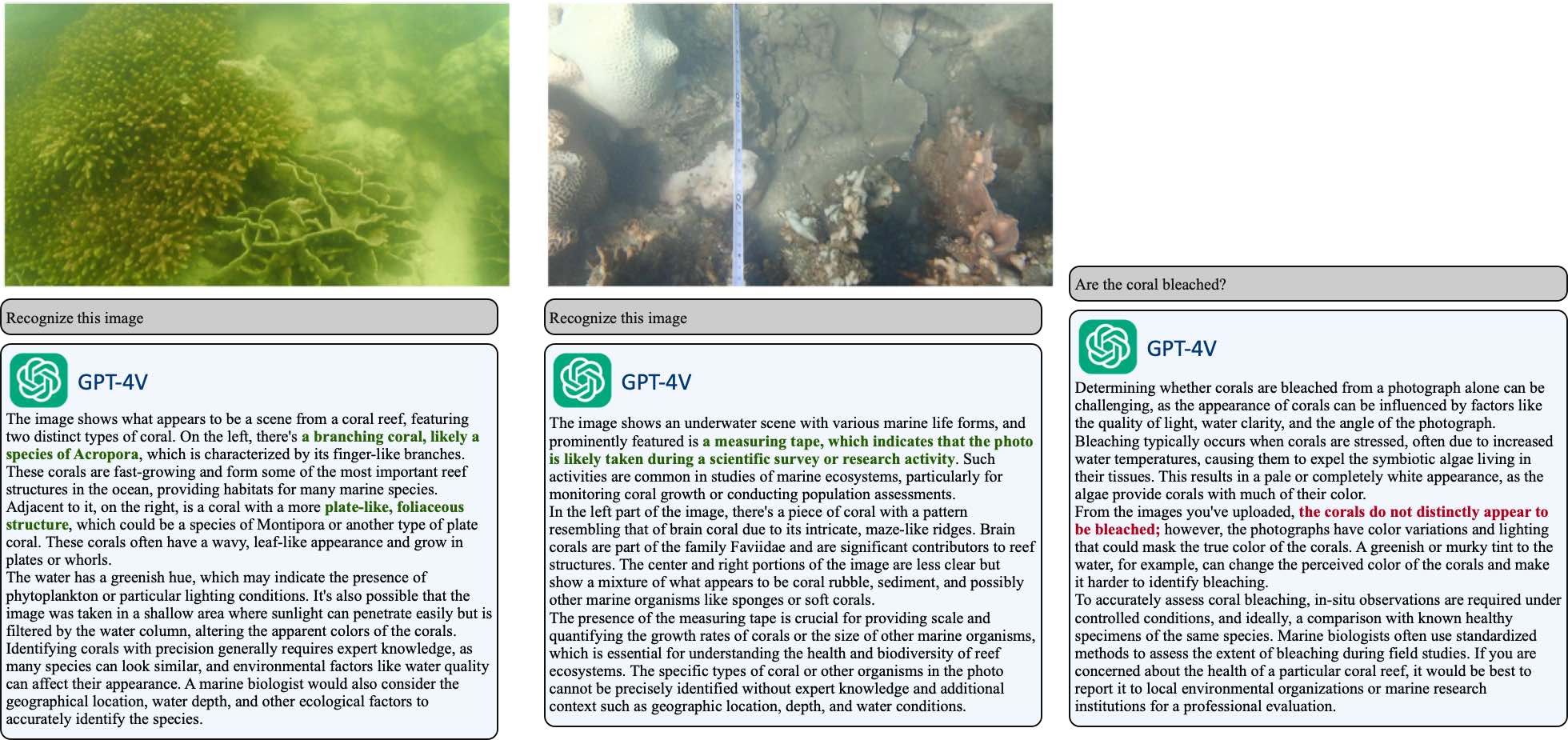}
\end{center}
\caption[Coral composition and coral bleaching detection]{Utilize GPT-4V for coral composition estimation and coral bleaching detection.}
\label{fig:coral_composition}
\end{figure}

We then examine the ability of GPT-4V to discriminate the coral reef composition from the visual images in Figure~\ref{fig:coral_composition}. GPT-4V could accurately recognize the coral reefs and missed the brain coral reefs. Moreover, we have also explored the ability of GPT-4V to understand the coral bleaching, which is linked to warming seas, can lead to declines in coral coverage. When being asked whether the coral reefs are bleached, GPT-4V has made a wrong judgment. GPT-4V cannot understand the meaning of ``bleaching'' and describes the degree of coral bleaching due to the lack of a reference color bar. 

\newpage
\subsubsection{Benthic Composition}
\begin{figure}[htbp]
\begin{center}
\includegraphics[width=\textwidth]{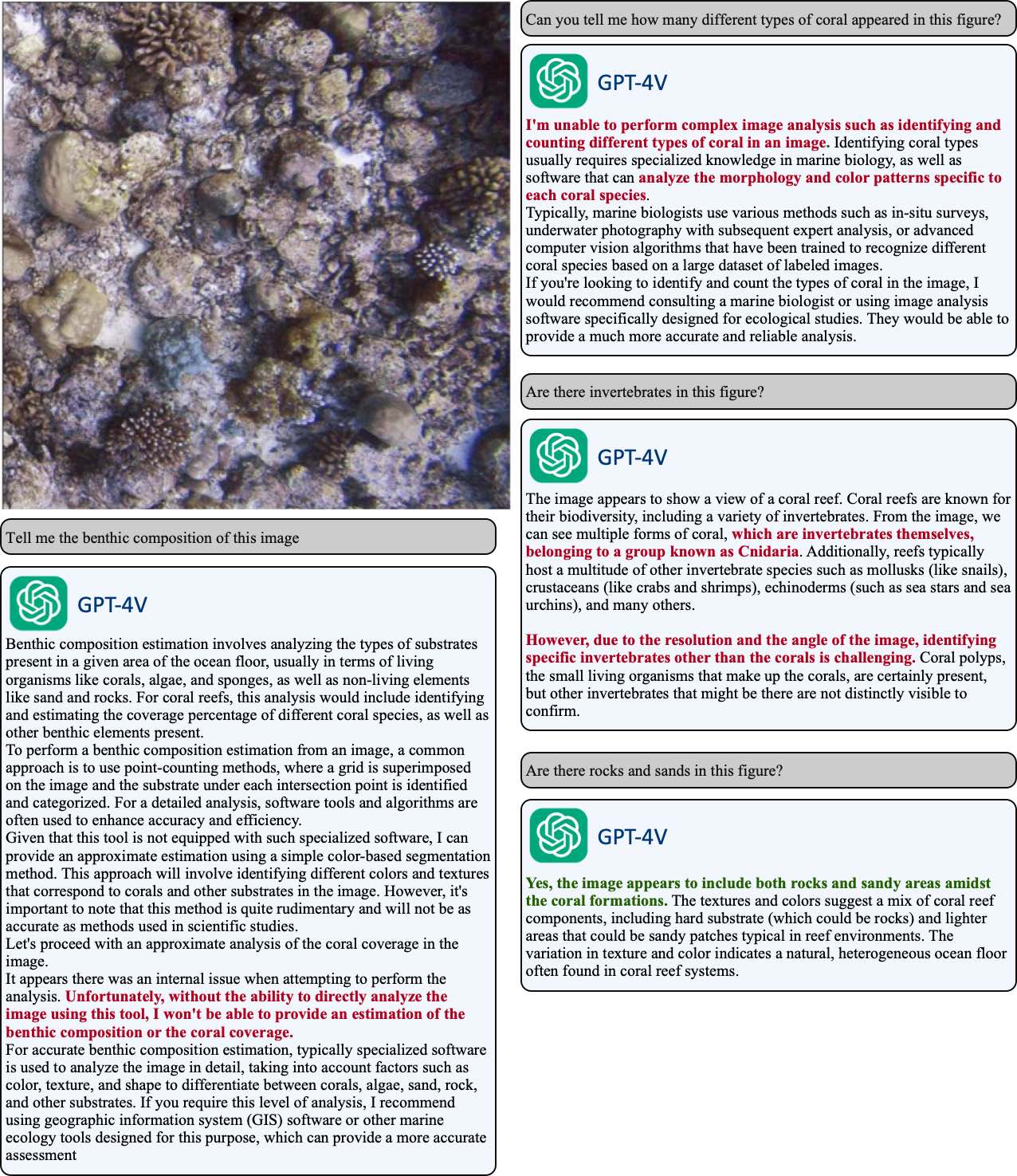}
\end{center}
\caption[Benthic composition estimation]{Utilize GPT-4V for the benthic composition estimation.}
\label{fig:benthic_composition}
\end{figure}
Understanding the benthic composition from the captured visual images could help researchers characterize and classify marine ecosystems based on the types of organisms and substrate present. Different benthic communities support distinct sets of species and play unique ecological roles. We explore the potential of utilizing GPT-4V to generate the benthic analysis data, which could be further used for monitoring the impact of factors like pollution, climate change, and habitat destruction. The results are illustrated in Figure~\ref{fig:benthic_composition}. We first ask GPT-4V to generate the benthic composition data (the composition of non-creatures and creatures) from the uploaded visual image and then identify how many types of coral reefs. Furthermore, we examine the ability of GPT-4V for benthic invertebrate identification (\emph{e.g.}, corals, sponges, mollusks, and worms), algae, and even certain fish species. 

Our experimental results show that GPT-4V nearly cannot achieve benthic composition statistics without utilizing an external professional analysis tool or being fed corresponding analysis data for final report generation. Even though GPT-4V could generate some very naive computer vision processing codes for analysis, the analyzed outputs are still very far from the requirement of a professional expert. Meanwhile, the whole processing and analysis procedure lacks the reasoning steps and support of the domain-specific evidence.

\newpage
\subsubsection{Relationship Summarization and Event Detection}
\begin{figure}[htbp]
\begin{center}
\includegraphics[width=\textwidth]{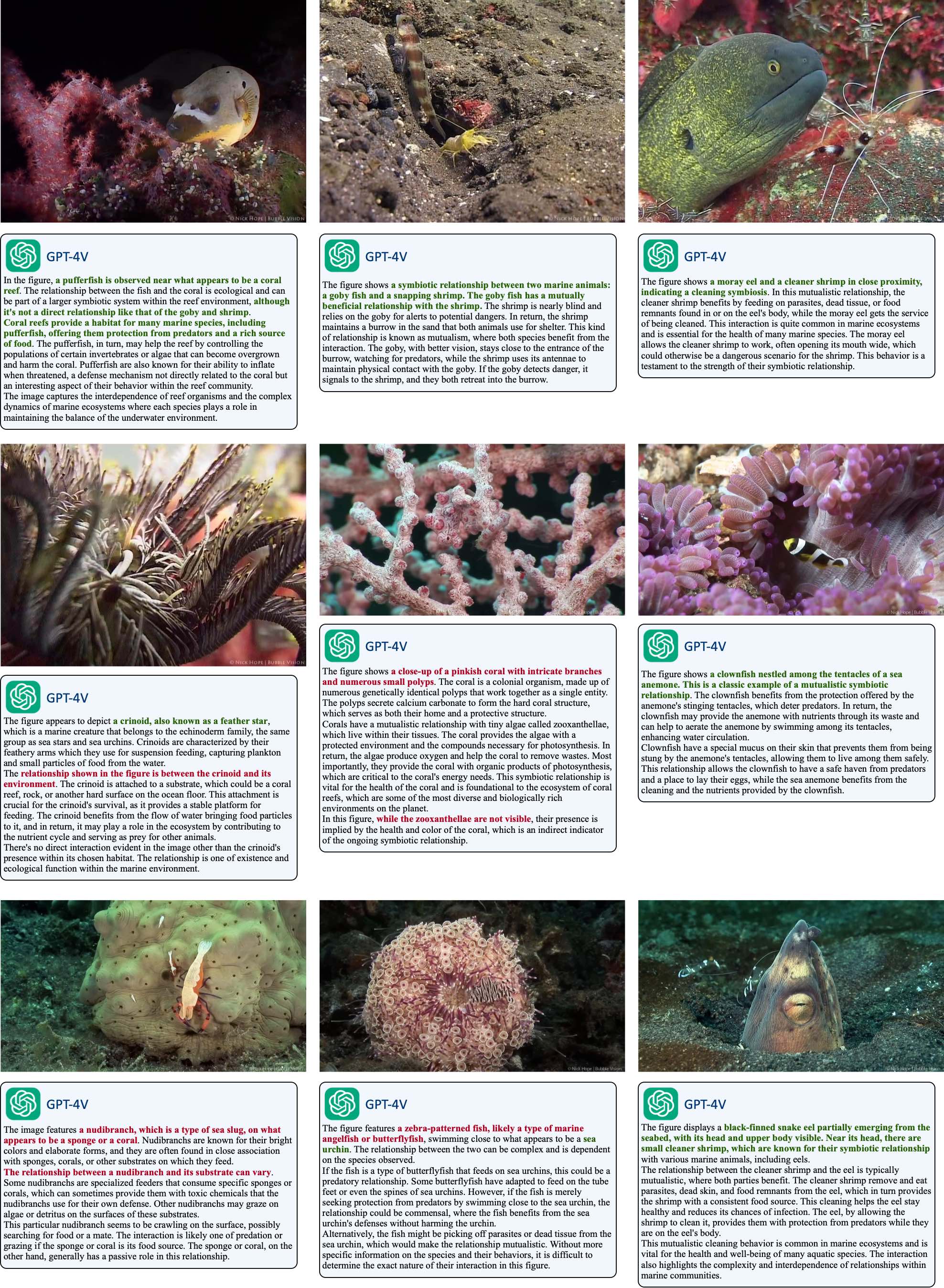}
\end{center}
\caption[Relationship summarization]{Utilize GPT-4V for relationship summarization from visual images. The prompts are ``Summarize the relationship between the objects within this figure''.}
\label{fig:relationship}
\end{figure}

\begin{figure}[htbp]
\begin{center}
\includegraphics[width=0.9\textwidth]{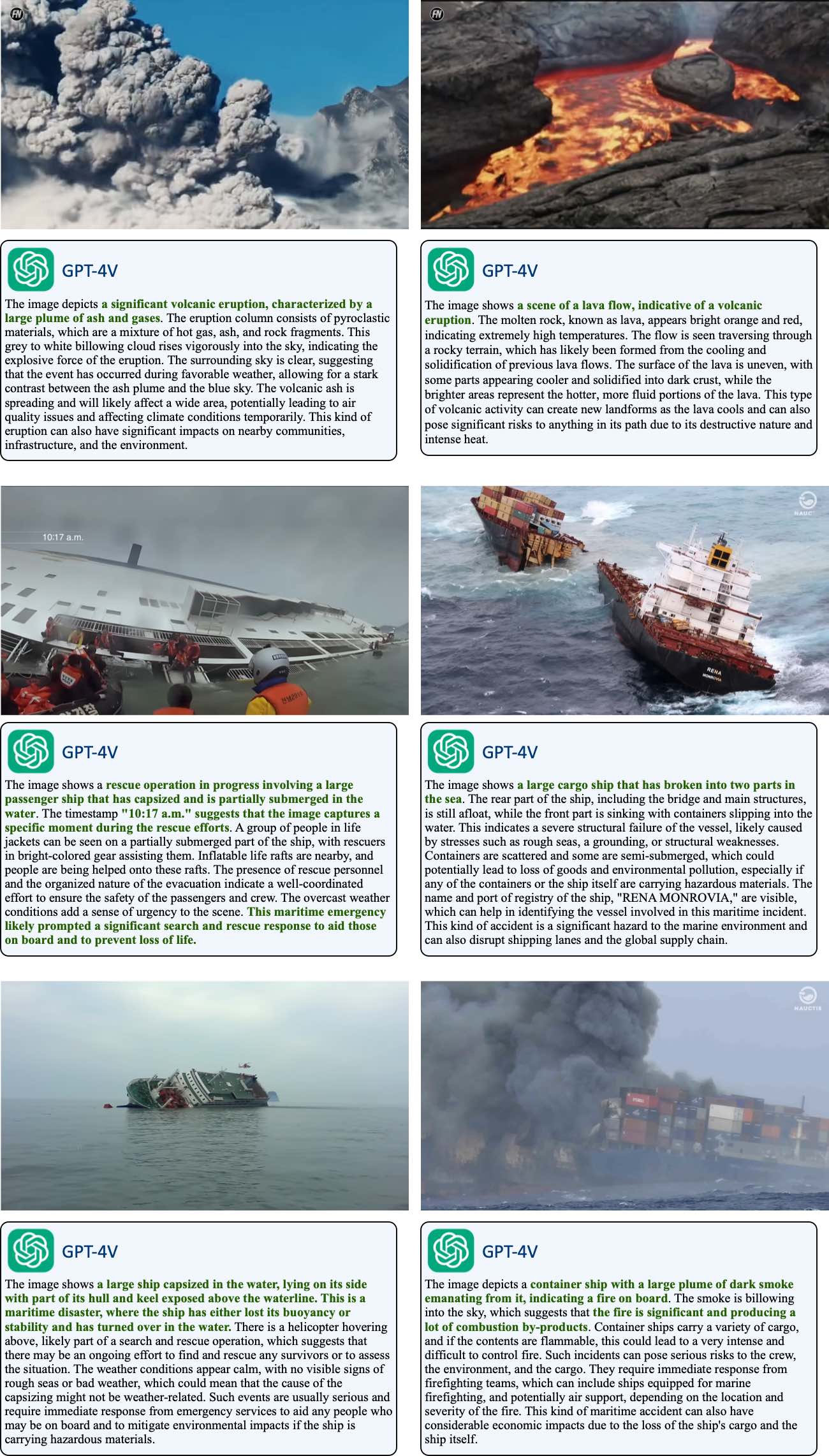}
\end{center}
\caption[Event detection]{Utilize GPT-4V for event detection. The prompts are ``Describe the event in this figure''.}
\label{fig:event}
\end{figure}

\noindent\textbf{Relationship summarization}. Exploring the relationships between marine creatures allows conservationists to make informed decisions about protecting vulnerable or endangered species. In this section, we assess the ability of GPT-4V to comprehend how different creatures interact and summarize the relationship between them, such as predator-prey relationships, symbiosis, competition, and mutualism. Such summarized marine relationships could gain insights into the behavior, evolution, and adaptation of species. It is worth noting that we mainly focus on the relationship summarization from the perspective of marine biology research. The qualitative results are reported in Figure~\ref{fig:relationship}. As demonstrated, GPT-4V has shown a satisfactory ability to understand and describe some well-known relationships between recognized objects, such as the symbiotic relationship between clownfish and the sea anemone. But in contrast, when GPT-4V fails to recognize the marine objects accurately, it will generate totally irrelative responses, and the responses are nearly based on its ``imagination''.

\noindent\textbf{Event detection}. Through event detection, domain experts could predict and mitigate the impacts of events like climate change and pollution. Some preliminary case studies about event detection are illustrated in Figure~\ref{fig:event}. We collect more samples about 1) identifying irregular behaviors, such as illegal fishing, vessel collisions, or suspicious activities, which can be crucial for maritime safety and security; 2) monitoring the changes of marine conditions, such as water levels, wave patterns, and coastal erosion; and 3) detecting abnormal events in marine images, which can help identify unusual events such as oil spills, coral bleaching, and marine pollution. Detecting these abnormalities early allows for a rapid response to mitigate environmental damage and protect marine ecosystems. The excitement of unveiling the unknown serves as a powerful motivator for researchers and explorers. From the early exploration as demonstrated in Figure~\ref{fig:event}, GPT-4V possesses a strong ability to understand the event presented in the visual images.

\newpage
\subsubsection{Framework and Flowchart Understanding}
\begin{figure}[htbp]
\begin{center}
\includegraphics[width=\textwidth]{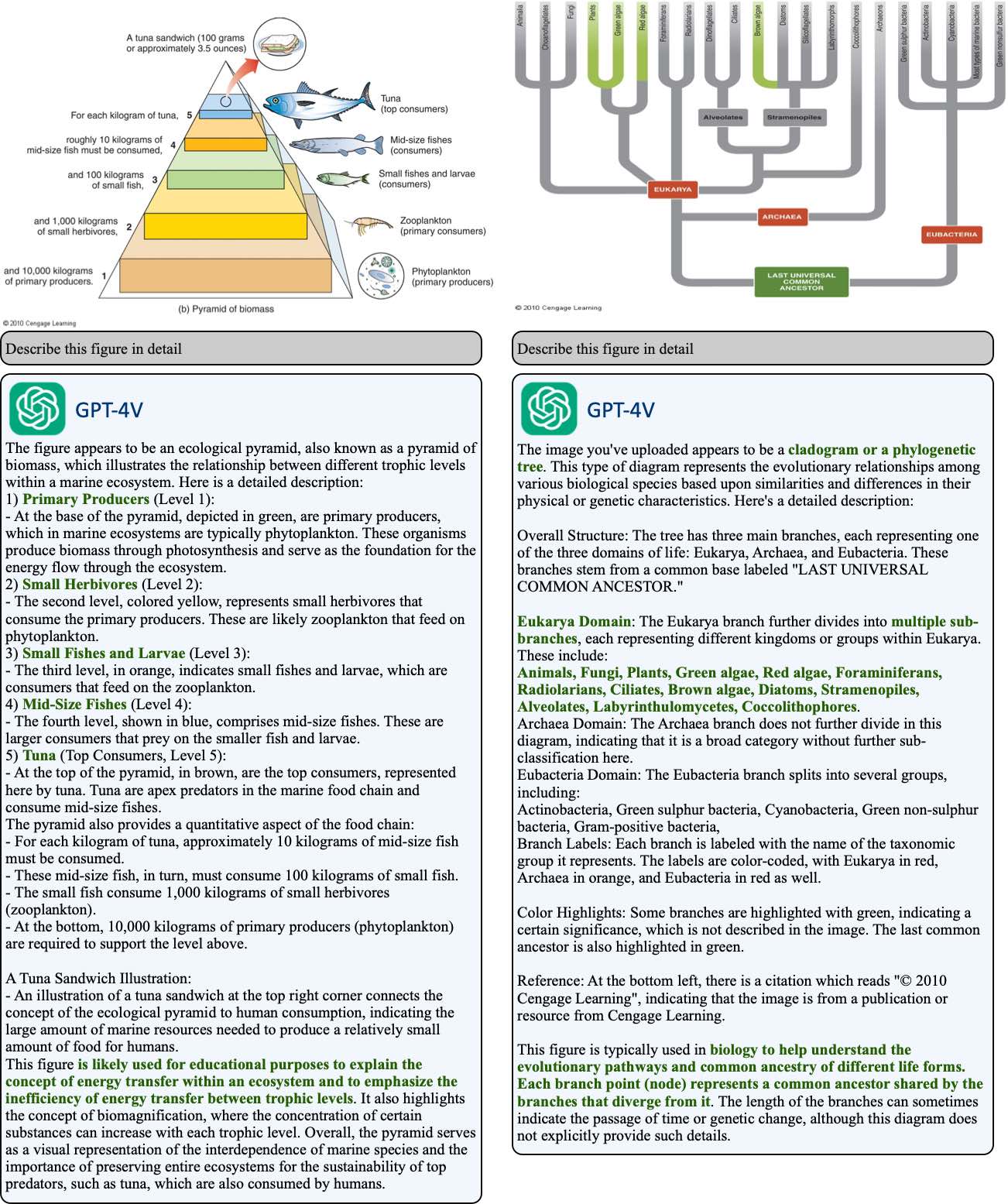}
\end{center}
\caption[Scientific figure understanding]{Utilize GPT-4V for scientific figure understanding in the marine research field.}
\label{fig:scientific}
\end{figure}

\begin{figure}[htbp]
\begin{center}
\includegraphics[width=\textwidth]{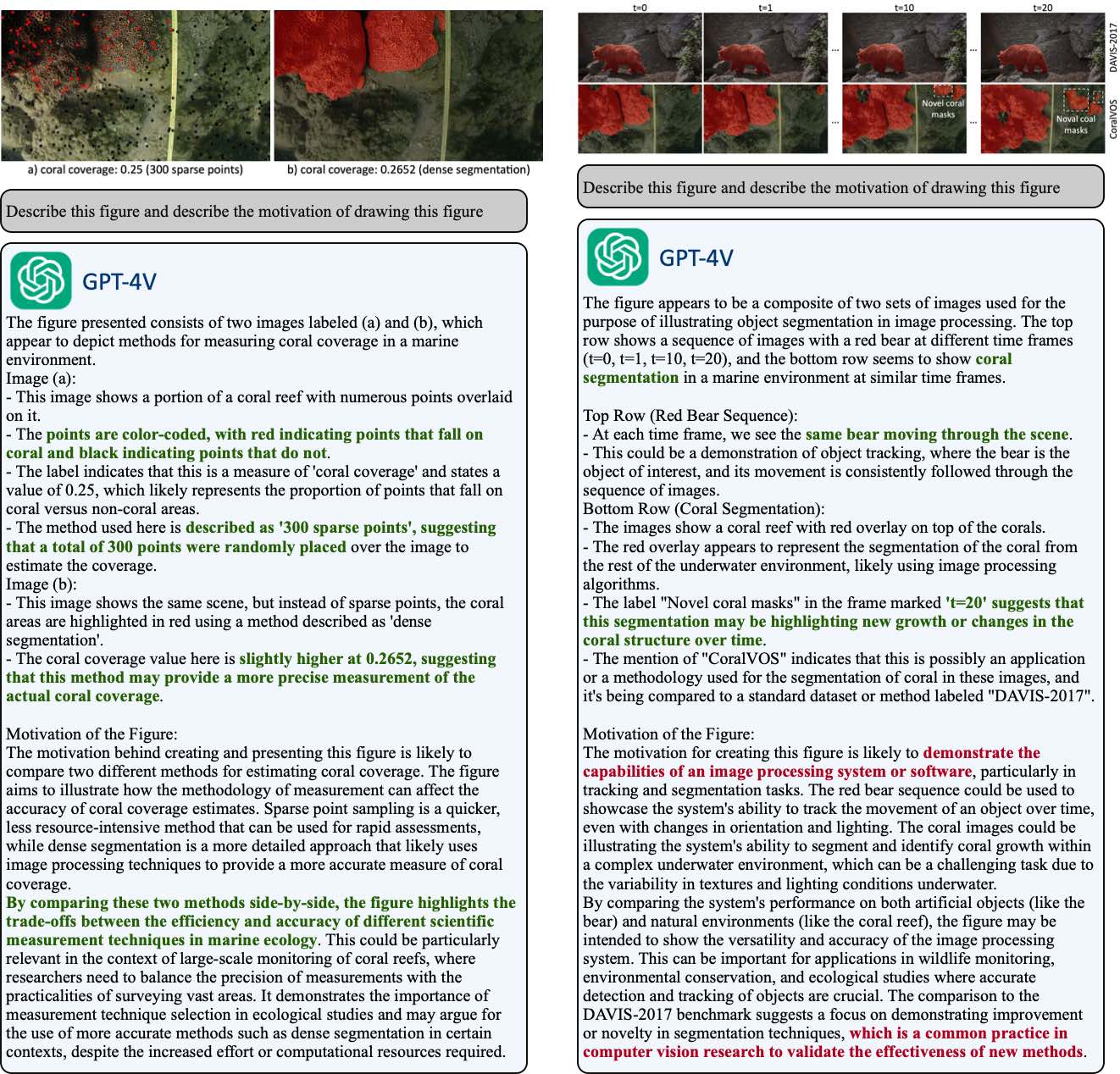}
\end{center}
\caption[Illustration figure understanding]{Utilize GPT-4V for illustration figure understanding. GPT-4V could understand the implicit intention.}
\label{fig:implicit_intention}
\end{figure}

\begin{figure}[htbp]
\begin{center}
\includegraphics[width=\textwidth]{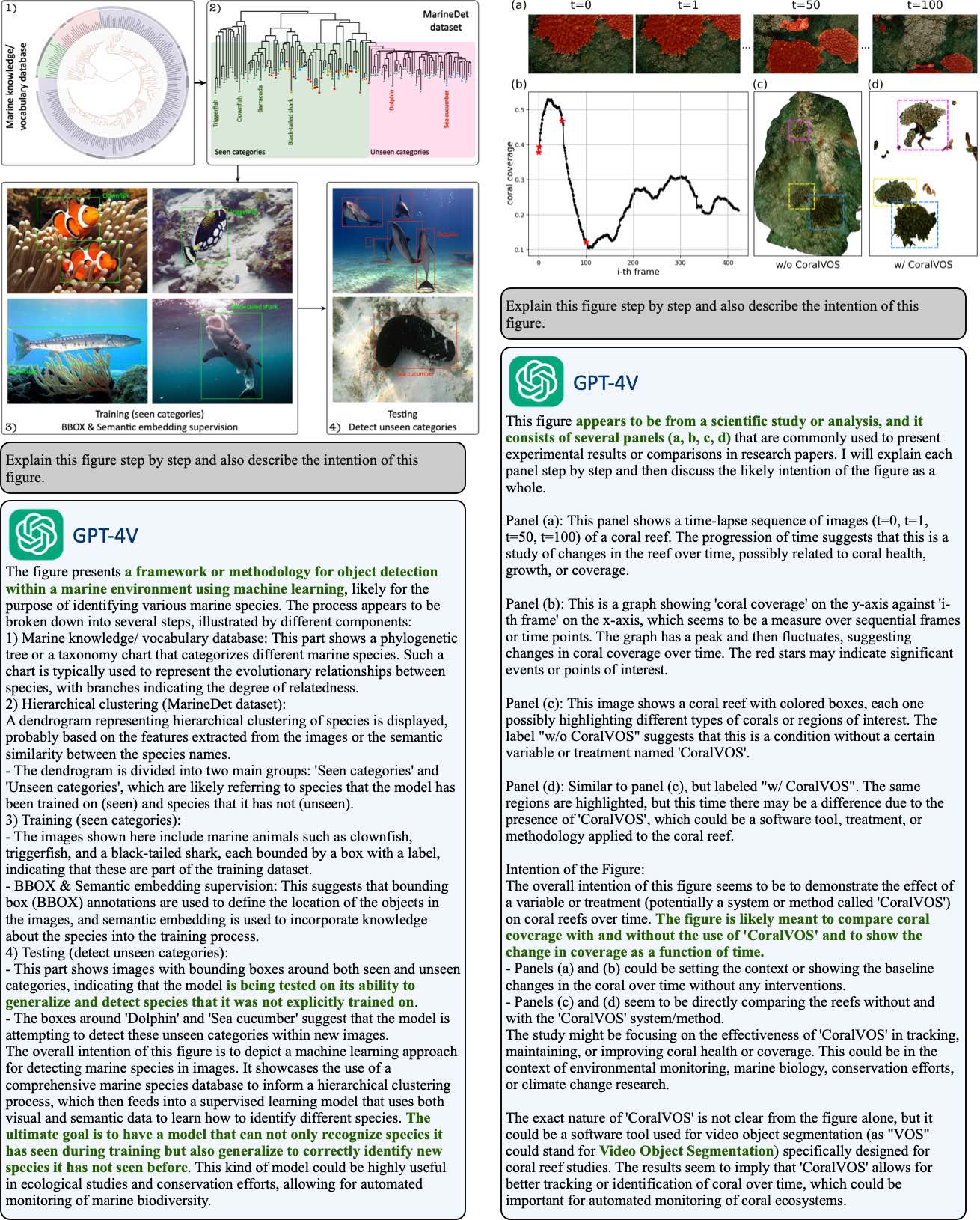}
\end{center}
\caption[Framework and flowchart understanding]{Utilize GPT-4V for framework and flowchart understanding. GPT-4V could explain the intermediate procedures step by step and explain the whole framework.}
\label{fig:framework_flowchart}
\end{figure}

We test whether GPT-4V showcases some detailed reasoning procedures and the ability to understand the inside intention of the designed images, including the framework and flow chart images. GPT-4V is required to explain the whole framework step by step and describe the intermediate step in detail. We provide visual reasoning results of GPT-4V from various fields in Figure~\ref{fig:scientific} (scientific figure understanding), Figure~\ref{fig:implicit_intention} (implicit intention understanding), and Figure~\ref{fig:framework_flowchart} (the framework understanding), respectively. Our exploration targets how GPT-4V understands and reasons for the high-level information from the figures as a whole.

As shown in Figure~\ref{fig:scientific}, GPT-4V has demonstrated a very strong OCR ability to extract text information from visual images. It could summarize the hierarchical relationship between different parts and extract the key elements of the whole figure. Besides, GPT-4V can understand the structure information and guess the source and usage of the uploaded scientific images. 

Furthermore, we observe that GPT-4V could understand the motivation of the illustration figures as demonstrated in Figure~\ref{fig:implicit_intention}. It could accurately describe the inside motivation of drawn figures. However, we have also observed the hallucination of GPT-4V. It will generate some information that does not exist within the image based on some extracted keywords (\emph{e.g.}, ``DAVIS-2017''). We attribute this phenomenon to the reason that GPT-4V may overfit its training data. How to prevent such hallucinations and alleviate the over-claim of GPT-4V is an important and valuable research direction.

Finally, we explore the ability of GPT-4V to understand and explain the framework or flowchart step by step in Figure~\ref{fig:framework_flowchart}. GPT-4V could accurately describe each part of the whole framework in detail and summarize the relationship between each part. Also, it demonstrates a satisfactory performance to understand the overall intention of the whole framework. 

\subsubsection{Aesthetic evaluation}
\begin{figure}[htbp]
\begin{center}
\includegraphics[width=\textwidth]{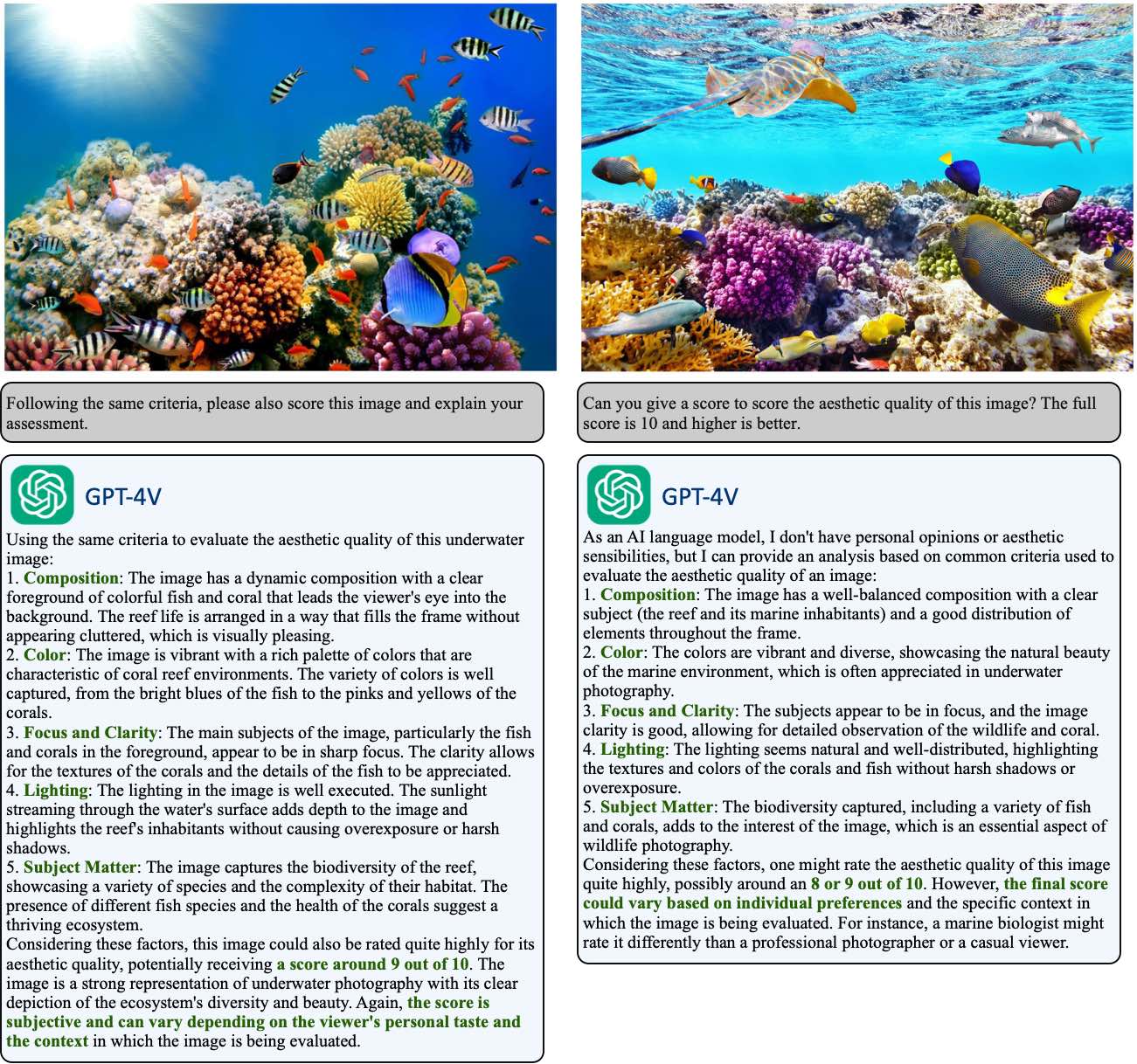}
\end{center}
\caption[Aesthetic evaluation]{Utilize GPT-4V for the aesthetic quality estimation. GPT-4V could explain the criteria of its assessments.}
\label{fig:aesthetic}
\end{figure}

We have also assessed the ability of GPT-4V to do the aesthetic evaluation. We manually constructed 50 marine images with high diversity then we uploaded the visual images to GPT-4V to generate the aesthetic score (scale of 10) based on the visual contents. To quantitatively evaluate the ability of GPT-4V for aesthetic assessment, we ask expert-level human labelers (3 annotators) to give the subjective scores towards the given marine images and we compute the mean value and the standard deviation. Then we first evaluate the alignment between the scores from GPT-4V and human labelers in terms of aesthetic measuring. We provide some qualitative results of GPT-4V in Figure~\ref{fig:aesthetic}. We observe that the scores generated by GPT-4V are highly correlated with human rating. GPT-4V successfully identifies the aesthetic quality of visual elements within the images and provides a comprehensive explanation for its scores. Our results reveal that GPT-4V achieves a promising agreement with humans on aesthetic quality assessment.

\subsubsection{Temporal Sequence Understanding} 

\begin{figure}[htbp]
\begin{center}
\includegraphics[width=\textwidth]{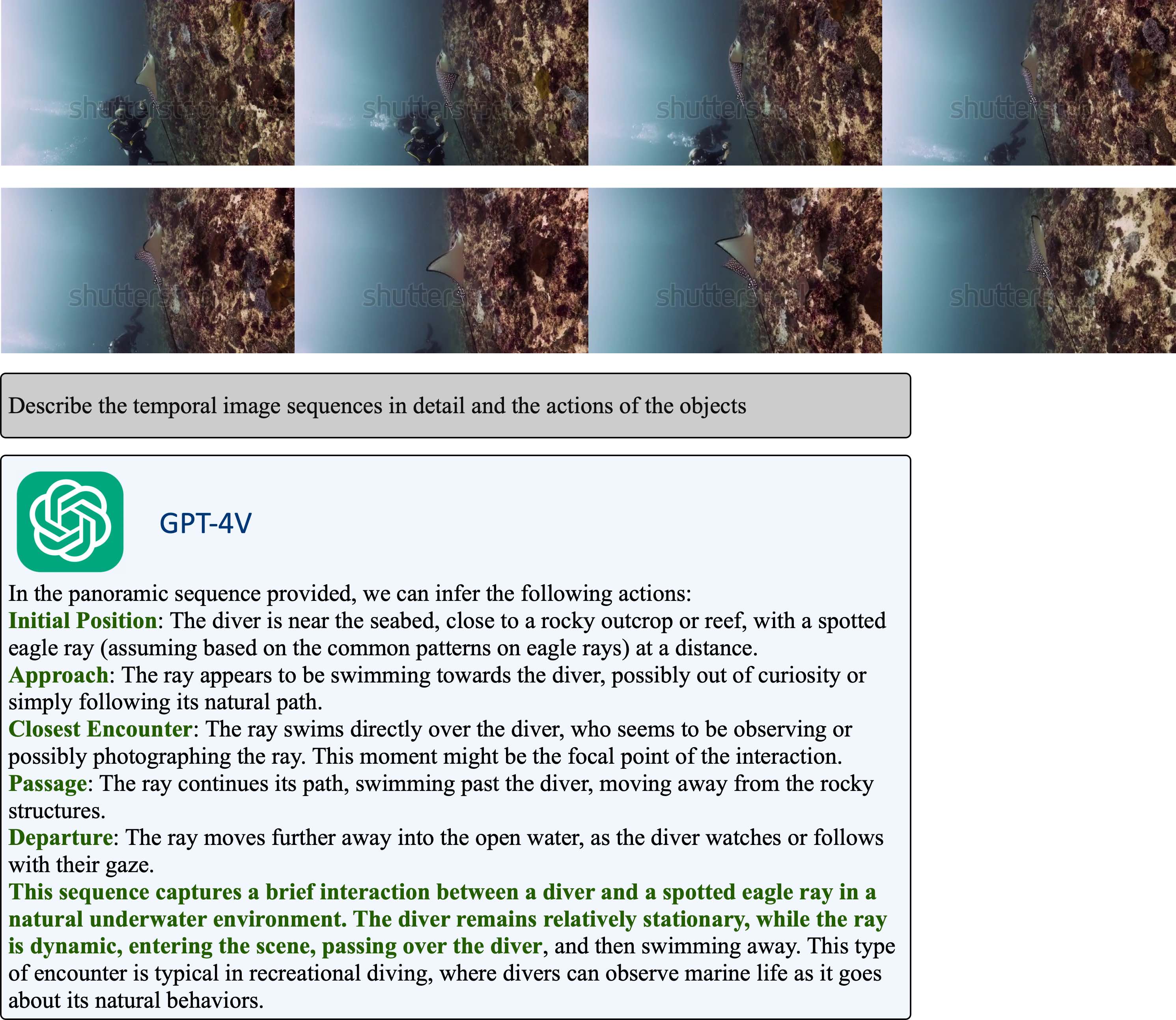}
\end{center}
\caption[Temporal content understanding]{Utilize GPT-4V for temporal content understanding from the video sequence.}
\label{fig:temporal}
\end{figure}

We finally explore the potential ability of GPT-4V for temporal sequence understanding. Given the consecutive image frames sampled from the video sequence (\emph{e.g.}, uniformly sampling 8 frames), we concatenate the sampled frames to one image and then ask GPT-4V to summarize the event that happened in the given video sequence. The temporal sequence understanding requires the MLLMs to fully comprehend the information within the visual sequence. Understanding the event of a marine clip could be very valuable for detecting the abnormal behavior of marine creatures and then preventing the potential disaster. The results are illustrated in Figure~\ref{fig:temporal}. As illustrated in Figure, GPT-4V demonstrates the capability to recognize the action in the images and provide a detailed description. It has shown a promising potential to understand scenes from video and visual story generation.

\subsection{Prompt Engineering}
In this section, we aim to explore the effectiveness of introducing the current prompt engineering techniques designed for general-purpose MLLMs for marine research. We mainly focus on three settings: 1) few-shot prompts; 2) self-consistency and 3) chain-of-thoughts. 

\begin{figure}[htbp]
\begin{center}
\includegraphics[width=\textwidth]{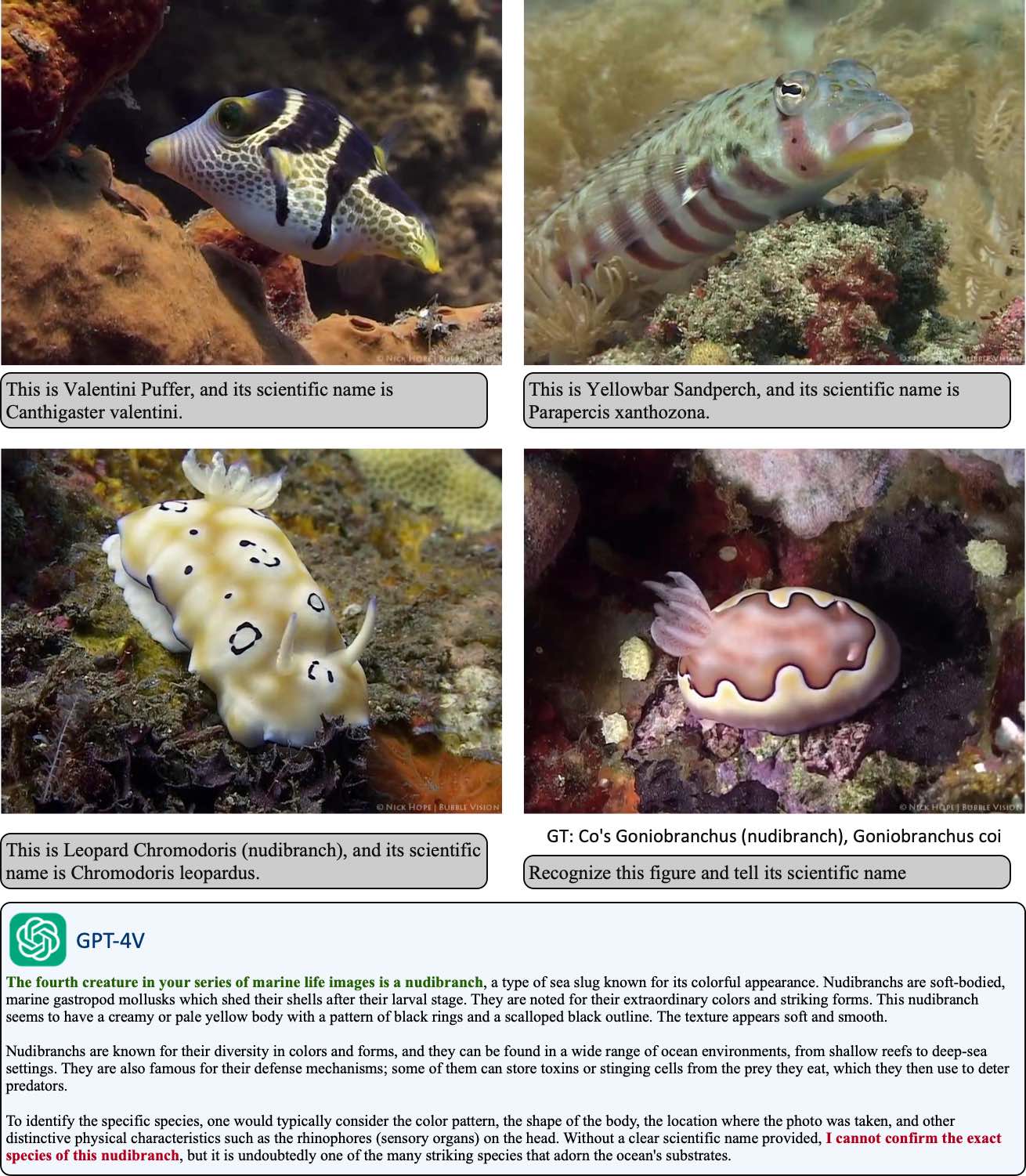}
\end{center}
\caption[Prompt engineering Case 1]{The marine object recognition results of GPT-4V with few-shot prompts provided.}
\label{fig:few-shot}
\end{figure}

Under the first setting, we feed the GPT-4V with few-shot samples with corresponding annotations to guide GPT-4V as a domain expert and help it better understand our questions. Then we ask the GPT-4V for a similar question as shown in Figure~\ref{fig:few-shot}. We observe that GPT-4V will still make mistakes and generate wrong responses even the few-shot prompts provided. We attribute this failure to the limited visual perception ability of GPT-4V. GPT-4V cannot effectively perform fine-grained object recognition. 

\begin{figure}[htbp]
\begin{center}
\includegraphics[width=\textwidth]{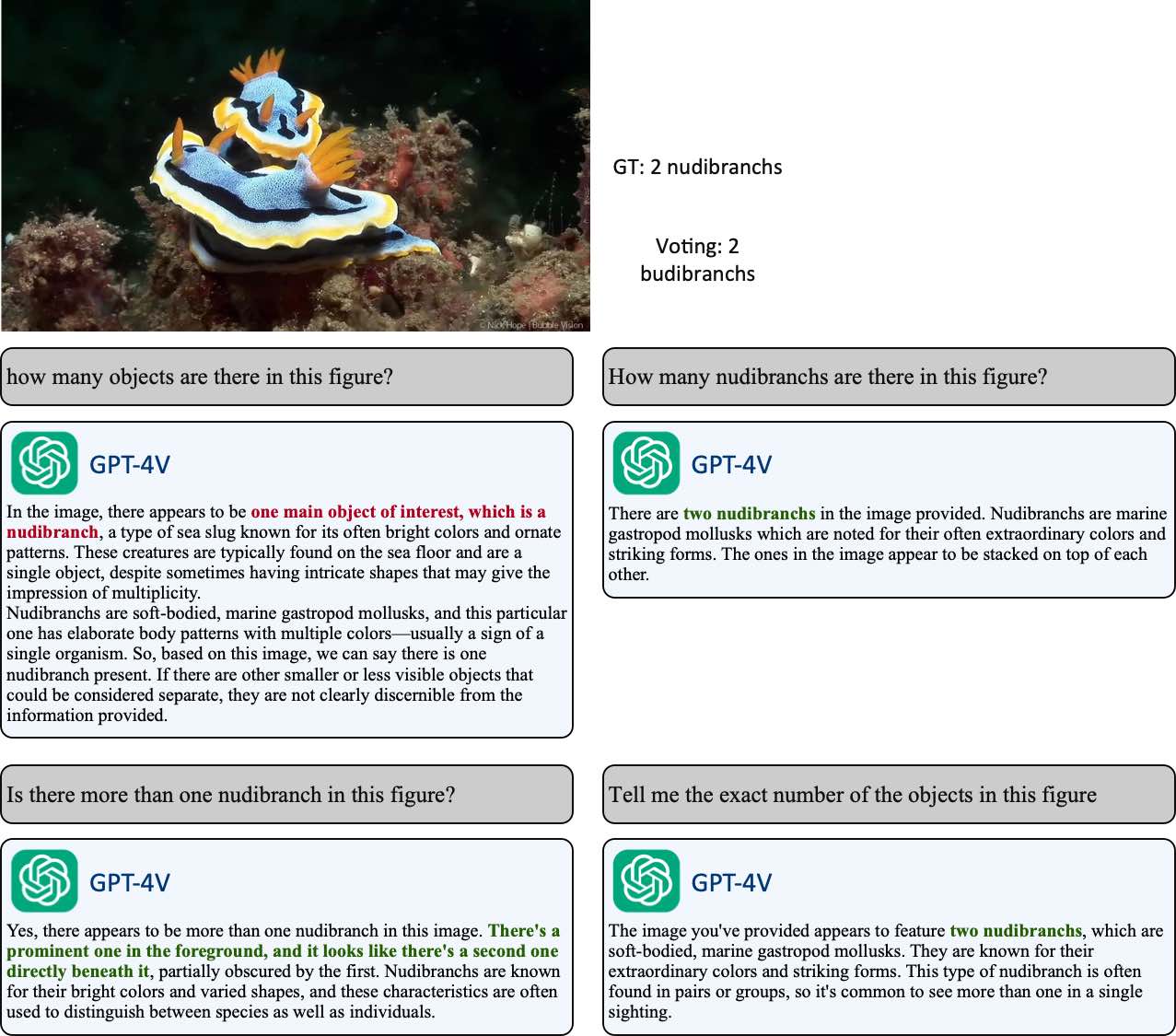}
\end{center}
\caption[Prompt engineering Case 2]{The self-consistency analysis of GPT-4V. Through voting, GPT-4V could generate more reliable responses.}
\label{fig:self-consistency}
\end{figure}

To explore the self-consistency of the GPT-4V, we ask the GPT-4V to do the object counting task based on various prompts and we then perform voting to get the final object count result. Through this, we aim to measure the self-consistency of GPT-4V for the same visual input and the robustness of its generated responses. Through voting or feeding GPT-4V with clearer prompts, GPT-4V could generate more reliable and accurate object counting results as demonstrated in Figure~\ref{fig:self-consistency}. 

\begin{figure}[htbp]
\begin{center}
\includegraphics[width=\textwidth]{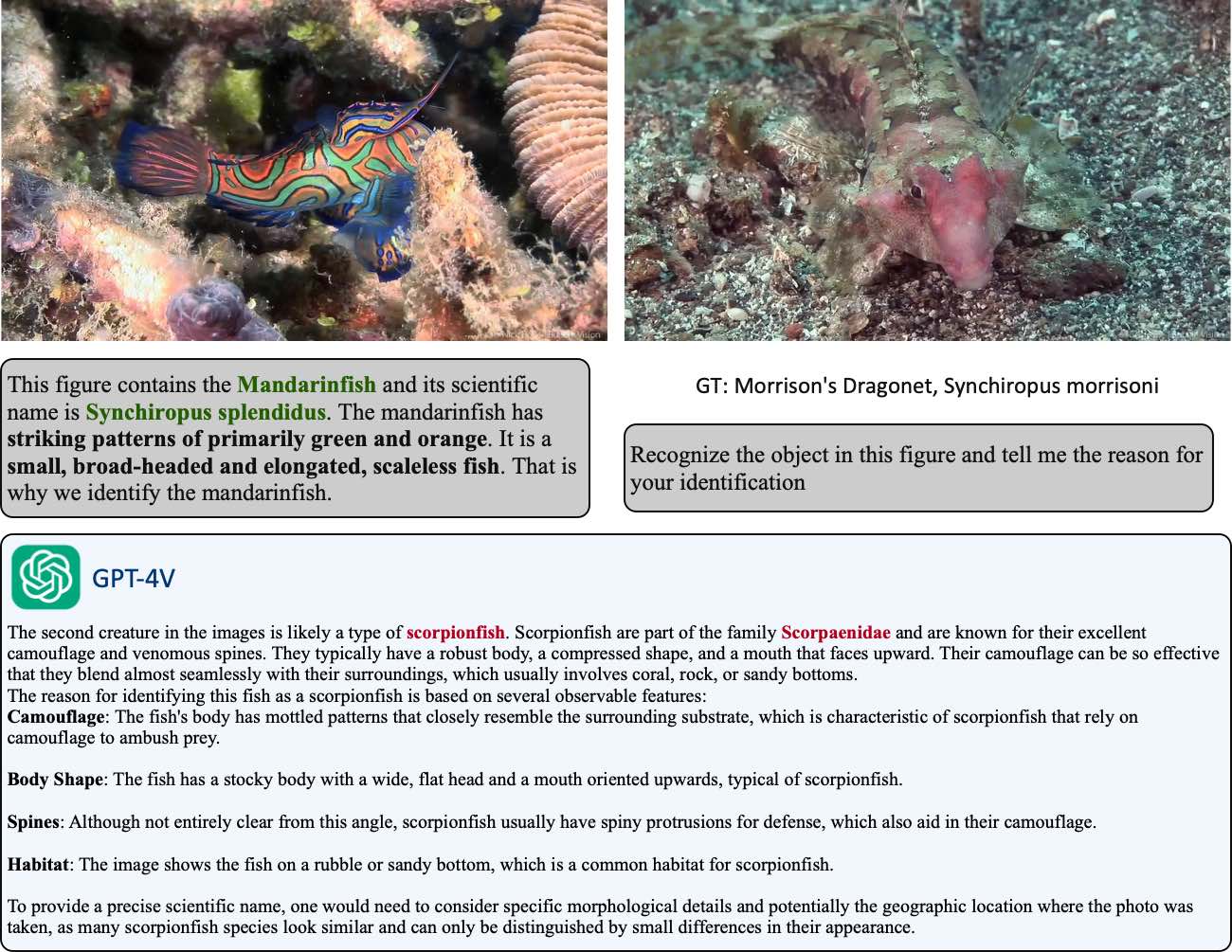}
\end{center}
\caption[Prompt engineering Case 3]{We prompt GPT-4V with detailed reasoning procedure and ask GPT-4V to explain its identification procedure.}
\label{fig:cot}
\end{figure}

Finally, we refer to the design of the chain-of-thoughts~\cite{yang2023dawn} and add some simple explanations in our input prompts. The GPT-4V is asked to follow our explanation procedure and understand the reasoning inside the recognition. In this way, GPT-4V could describe more about its judgment and illustrate more supporting evidence. The results are reported in Figure~\ref{fig:cot}. We observe that GPT-4V sucks the ability to accurately recognize marine objects even GPT-4V could generate plausible and detailed descriptions about the wrongly recognized object. 

To sum up, the current prompt engineering techniques cannot heavily promote the visual recognition ability of GPT-4V on marine images. GPT-4V will still make mistakes for fine-grained marine object recognition and prompt engineering cannot alleviate the hallucination issue, effectively. To address these issues, more training data from the marine field should be included for further promoting the recognition ability of GPT-4V. 

\newpage
\section{Discussions and Future Directions}
\subsection{Discussions}
\noindent\textbf{Possible for educational tool?}
While the performance of the GPT-4V is promising, we ask whether GPT-4V could be viewed as a potential educational tool that may in the future augment, but not replace, the nuanced analysis provided by trained marine professionals. GPT-4V could also play as a pivotal role in fostering a deeper understanding and appreciation for marine life among users of all ages and backgrounds. Through our findings in this study, we conclude that GPT-4V is far from generating valuable insights for domain experts. 

\noindent\textbf{Possible for labeling tool?} With easy access to GPT-4V, it could actively encourage citizen science participation as a labeling tool, transforming ordinary individuals into valuable contributors to marine research. From our findings, we observe that GPT-4V cannot serve as a labeling tool for a wide spectrum of marine images since GPT-4V still makes many mistakes for challenging images. Moreover, such labeling is also only limited to image-level scene understanding. GPT-4V cannot generate accurate descriptions for the very fine-grained details. 

\noindent\textbf{Sample Bias}. In our study, the testing samples are manually constructed, inevitably incorporating individual preferences and subjectivity. More importantly, our involved testing samples may not comprehensively represent real-world cases, and potentially over-estimate or down-estimate the challenges of utilizing GPT-4V for marine analysis.

\subsection{Future Works}
Our findings emphasize the need for continued research to enhance the accuracy and expertise of responses generated by GPT-4V. We hope that this study can inspire more comprehensive and targeted research into utilizing multimodal systems such as GPT-4V for domain-specific research and analysis. By harnessing the capabilities of these models, we can better meet the professional demands of experts, ultimately including the domain experts in the major users of GPT-4V. Furthermore, based on the feedback and further prompts from the domain experts, a fundamental question arises, could GPT-4V revise its responses over time? Such feedback-driven MLLM would further promote the user experience for obtaining more precise responses.

Through our experimental results, we have observed that GPT-4V cannot achieve fine-grained and accurate marine object recognition to satisfy the requirements of the domain experts. More training data from the marine field should be included to promote the visual recognition ability of GPT-4V. Furthermore, we also demonstrate that GPT-4V has shown a very limited ability to handle advanced marine analysis (\emph{e.g.}, counting, coverage estimation, composition statistic, \textit{etc}) without utilizing an external professional tool. More domain-specific instruction-following data should be constructed to help GPT-4V yield explicit intermediate analysis results. 

\section{Conclusion}
In this paper, our investigation of GPT-4V on marine analysis demonstrates some valuable findings and insights of MLLMs concerning visual understanding, logical reasoning, and expert capacity, indicating that there remains a considerable distance toward strong artificial intelligence as a domain expert. 

\newpage
\bibliography{ref}
\bibliographystyle{iclr2023_conference}
\newpage
\end{document}

%% file: math_commands.tex
\usepackage{amsmath,amsfonts,bm}

\def\eqref#1{equation~\ref{#1}}

\def\1{\bm{1}}

\DeclareMathAlphabet{\mathsfit}{\encodingdefault}{\sfdefault}{m}{sl}
\SetMathAlphabet{\mathsfit}{bold}{\encodingdefault}{\sfdefault}{bx}{n}

